\pdfoutput=1
\documentclass{article}

\newif\ificml
\icmlfalse

\usepackage{lib/arxiv}
\usepackage{amsmath}

\usepackage{amsmath,amsfonts,amssymb,amsthm,bm}
\usepackage{bbold}

\def\eqref#1{equation~\ref{#1}}

\def\1{\bm{1}}

\def\rmA{{\mathbf{A}}}
\def\rmB{{\mathbf{B}}}
\def\rmC{{\mathbf{C}}}

\def\rmK{{\mathbf{K}}}

\def\rmS{{\mathbf{S}}}

\def\rmU{{\mathbf{U}}}

\def\rmW{{\mathbf{W}}}

\def\va{{\bm{a}}}
\def\vb{{\bm{b}}}
\def\vc{{\bm{c}}}
\def\vd{{\bm{d}}}
\def\ve{{\bm{e}}}
\def\vf{{\bm{f}}}

\def\vh{{\bm{h}}}
\def\vi{{\bm{i}}}

\def\vk{{\bm{k}}}
\def\vl{{\bm{l}}}

\def\vo{{\bm{o}}}

\def\vq{{\bm{q}}}
\def\vr{{\bm{r}}}

\def\vu{{\bm{u}}}
\def\vv{{\bm{v}}}
\def\vw{{\bm{w}}}
\def\vx{{\bm{x}}}
\def\vy{{\bm{y}}}
\def\vz{{\bm{z}}}

\def\mA{{\bm{A}}}
\def\mB{{\bm{B}}}

\def\mW{{\bm{W}}}

\DeclareMathAlphabet{\mathsfit}{\encodingdefault}{\sfdefault}{m}{sl}
\SetMathAlphabet{\mathsfit}{bold}{\encodingdefault}{\sfdefault}{bx}{n}

\def\gS{{\mathcal{S}}}

\newcommand{\R}{\mathbb{R}}

\newcommand{\softmax}{\mathrm{softmax}}
\newcommand{\sigmoid}{\sigma}

\usepackage{tikz}

\definecolor{tfcolor}{HTML}{2f4b7c}
\definecolor{tf1color}{HTML}{61b2e5}
\definecolor{tf2color}{HTML}{5397cb}
\definecolor{tf4color}{HTML}{477db1}
\definecolor{tf8color}{HTML}{3b6396}
\definecolor{tf12color}{HTML}{2f4b7c}
\definecolor{mambacolor}{HTML}{3b9247}
\definecolor{glacolor}{HTML}{6fe77d}
\definecolor{retnetcolor}{HTML}{6b6395}
\definecolor{ltfcolor}{HTML}{998dbe}
\definecolor{lstmcolor}{HTML}{aeaeae}
\definecolor{hyenacolor}{HTML}{8b3800}
\definecolor{htcolor}{HTML}{c76c01}
\definecolor{rwkvcolor}{HTML}{ffa600}
\definecolor{sfcolor}{HTML}{ff9984}

\newcommand{\tikzcircle}[2][red,fill=red]{\tikz[baseline=-0.5ex]\draw[#1,radius=#2] (0,0) circle ;}

\definecolor{modelboxcolor}{rgb}{0.9,0.9,0.9}
\definecolor{modelshapefill}{rgb}{0.98,0.98,0.98}

\newcommand{\tf}{{Transformer}}

\newcommand{\rwkv}{{RWKV}}

\newcommand{\lstm}{{LSTM}}

\newcommand{\sfour}{{S4}}

\newcommand{\hyena}{{Hyena}}

\newcommand{\hthree}{{H3}}

\usepackage[utf8]{inputenc} %
\usepackage[T1]{fontenc}    %
\usepackage{hyperref}       %
\usepackage{url}            %
\usepackage{subfig}
\usepackage{booktabs}
\usepackage{amsfonts}       %
\usepackage{nicefrac}       %
\usepackage{microtype}      %
\usepackage{lipsum}         %
\usepackage{graphicx}
\usepackage[comma,authoryear,round]{natbib}
\usepackage{doi}

\usepackage{amssymb}
\usepackage{float}
\usepackage{multicol}
\usepackage{xcolor}
\usepackage{float}
\usepackage{titlesec}

\usepackage{xspace}
\usepackage{pmboxdraw}
\usepackage{adjustbox}
\usepackage[capitalize, noabbrev]{cleveref}
\usepackage{cuted}
\usepackage[most]{tcolorbox}
\usepackage[shortlabels,inline]{enumitem}
\usepackage{pythonhighlight} 

\usepackage{subcaption}

\newcommand\boxRight{\textSFii\pmboxdrawuni{2574}}

\newcommand\boxSpace{\hspace{1.5em}}
\newcommand{\septoken}{{\ \tikzcircle[fill=black]{1pt}\ }}

\hypersetup{
    colorlinks,
    linkcolor={red!50!black},
    citecolor={blue!80!black},
    urlcolor={blue!80!black}
}
\newenvironment{rcases}   
{\left.\begin{aligned}}  
 {\end{aligned}\right\rbrace}  

\definecolor{brandblue}{rgb}{0.34, 0.7, 1}

\tcbset {
  base/.style={
    arc=0.0mm, 
    bottomtitle=0mm,
    boxrule=0mm,
    colbacktitle=black!10!white, %
    coltitle=black, 
    colback=white,
    left=2.5mm,
    leftrule=0.4mm,
    bottomrule=0.0mm,
    right=3.5mm,
    title={#1},
    toptitle=0.75mm, 
  }
}

\newtcolorbox{mainbox}[1]{
  colframe=tfcolor, %
  base={#1}
}

\newcommand\ourdataset{\textsc{RegBench}\xspace}
\usepackage{algorithm}
\usepackage{algorithmic}

\date{}

\usepackage{authblk}

\newif\ifuniqueAffiliation
\uniqueAffiliationtrue

\ifuniqueAffiliation %
\author{{Ekin Aky\"urek} \hspace{8mm} {Bailin Wang} \hspace{8mm}  {Yoon Kim}  \hspace{8mm} {Jacob Andreas} \\ 
MIT CSAIL \\\vspace{0.1em}
\texttt{\{akyurek, bailinw, yoonkim, jda\}@mit.edu}
}
\else

\affil[1]{CSAIL}
\affil[2]{CSAIL}
\fi

\def\balpha{{\bm{\alpha}}}
\def\bbeta{{\bm{\beta}}}

\providecommand{\swish}{\text{swish}}

\def\vkv{{\bm{kv}}}

\renewcommand{\headeright}{}
\renewcommand{\undertitle}{}
\title{In-Context Language Learning:\\ Architectures and Algorithms}
\renewcommand{\shorttitle}{In-Context Language Learning: Architectures and Algorithms}
\hypersetup{
pdftitle={In-Context Language Learning: Architectures and Algorithms},
pdfsubject={machine learning, nlp, deep neural networks},
pdfauthor={Ekin Aky\"urek, Bailin Wang, Yoon Kim, Jacob Andreas},
pdfkeywords={in-context learning, language modeling},
}

\begin{document}
\maketitle
\begin{abstract}
Large-scale neural language models (LMs) exhibit a remarkable capacity for \emph{in-context learning (ICL)}: they can infer novel functions from datasets provided as input. Most of our current understanding of when and how ICL arises comes from LMs trained on extremely simple learning problems like linear regression and associative recall. There remains a significant gap between these model problems and the ``real'' ICL exhibited by LMs trained on large text corpora, which involves not just retrieval and function approximation but free-form generation of language and other structured outputs.  In this paper, we study ICL through the lens of a new family of model problems we term \textbf{in context language learning (ICLL)}. In ICLL, LMs are presented with a set of strings from a formal language, and must generate additional strings from the same language. ICLL is designed to be simple enough to study in small-scale LMs, but complex enough to capture the key features of ICL in large-scale LMs. Here we focus on in-context learning of regular languages generated by \textbf{random finite automata}. We evaluate a diverse set of neural sequence models (including several RNNs, Transformers, and state-space model variants) on regular ICLL tasks, aiming to answer three questions: (1) \emph{Which model classes} are empirically capable of ICLL? (2) \emph{What algorithmic solutions} do successful models implement to perform ICLL? (3) \emph{What architectural changes} can improve ICLL in less performant models? We first show that Transformers significantly outperform neural sequence models with recurrent or convolutional representations on ICLL tasks. Next, we provide evidence that their ability to do so relies on specialized ``n-gram heads'' (higher-order variants of previously-described ``induction heads'') that compute input-conditional next-token distributions. Finally, we show that hard-wiring these heads into Transformer, recurrent and convolutional models improves performance not just on synthetic ICLL, but natural language modeling---reducing the perplexity of 340M-parameter models by up to 1.14 points (6.7\%) on the SlimPajama dataset. Our results highlight the usefulness of in-context formal language learning as a tool for  understanding ICL   in models of natural text.

\end{abstract}
\ificml
\else
\begin{figure}[h]
    \centering    \includegraphics[height=0.30\textwidth]{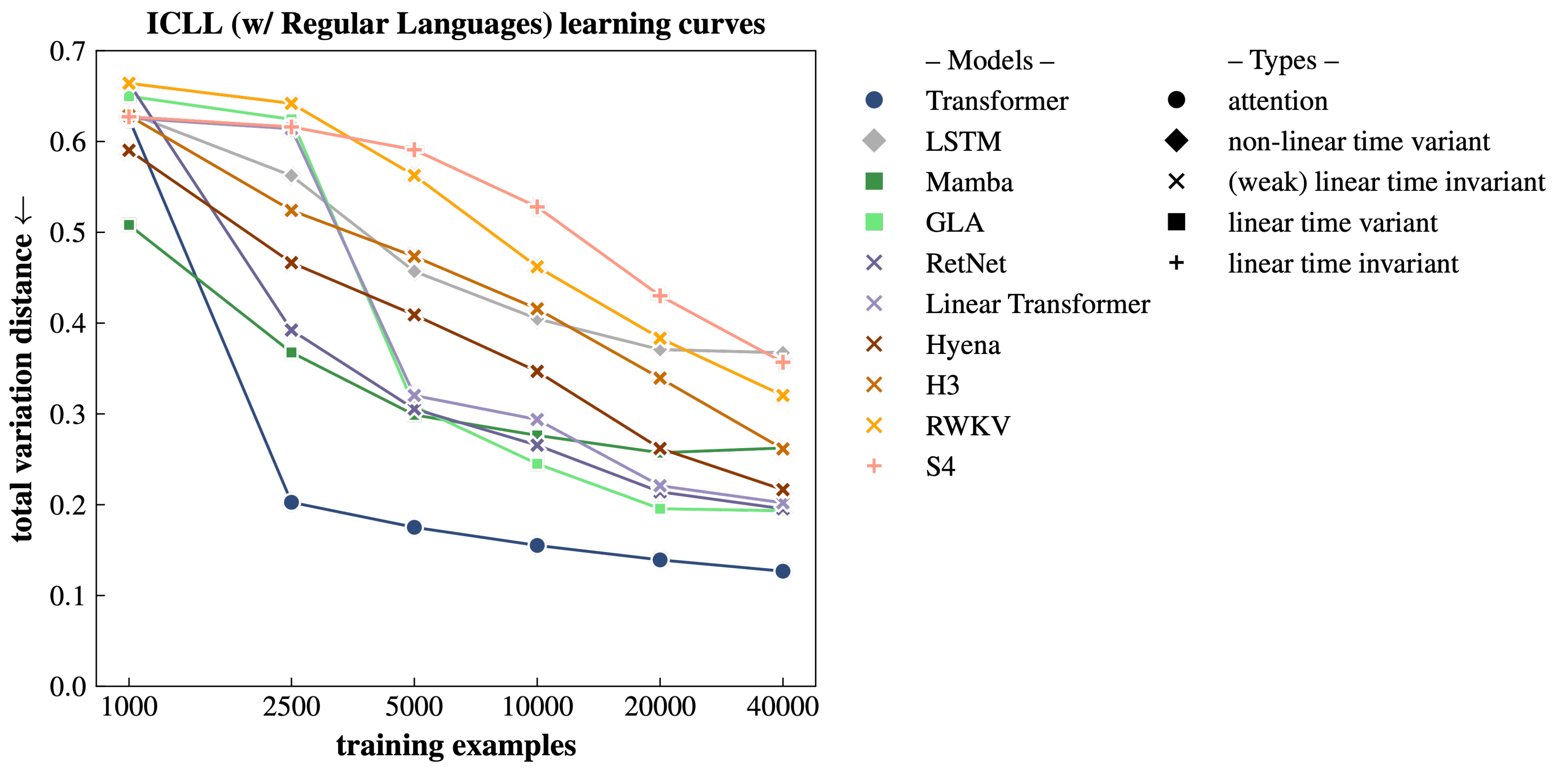}
    \caption{\textbf{Transformers effectively learn to perform in-context language learning}, but other model classes (including LSTMs and especially models with time invariant transitions---i.e. efficient convolutional forms) struggle to do so.}
    \label{fig:teaser}
\end{figure}
\fi
\section{Introduction}
\label{sec:intro}
One of the most striking features of modern neural language models is their capacity for \textbf{in-context learning} (ICL)---the ability to infer a conditional or unconditional distribution over natural language strings simply by performing next-token prediction following a sequence of examples from the distribution of interest. 
ICL is a crucial tool for steering large pre-trained language models (LMs), and 
a growing body of work aims to understand \emph{when} and \emph{how} these LMs perform ICL. Because of the complexity of large-scale LMs trained on natural language text (as well as the lack of public information about many LMs' training data), almost all work on understanding ICL has instead focused on smaller LMs trained on simple \textbf{model problems} like in-context linear regression \citep{garg2022can}, character classification \citep{chan2022data}, and associative recall \citep{dao2022hungry}. Despite their simplicity, these model problems have played a key role in identifying properties (and limitations) of ICL in current LMs.

However, there remains a significant gap between these model problems and the kinds of capabilities exhibited by LMs trained on large datasets of natural language text. In particular, most model problems require relatively simple forms of learning: computing a fixed function of the entire training set \citep{akyurek2022learning, von2023transformers, von2023uncovering}, or retrieving a single example relevant to the current input \citep{dao2022hungry}. In contrast, natural LMs exhibit richer and much more varied forms of ICL---in some cases producing structured generative models of text or code based on a handful of inputs \citep{shin-van-durme-2022-shot,drozdov2022compositional}.
This requires not just extracting a single summary statistic or example, but instead modeling multiple conditional distributions and re-composing fragments of training examples.

How can we systematically study these more complex forms of ICL? In this paper, we introduce a new family of model ICL problems that we collectively term \textbf{in-context language learning (ICLL)}. In ICLL, LMs are prompted with a finite collection of strings from an unknown formal language, and must infer the distribution over strings corresponding to the full language (\cref{fig:icllbench}). ICLL exercises essential features of ICL in natural models: it features structured outputs, probabilistic predictions, and compositional reasoning about input data.
In this paper, we present a focused study of ICLL in \textbf{regular languages}---the class of formal languages generated by finite automata. During ICLL training, each training \emph{example} consists of a sequence of strings from the same language. Different examples feature strings from different languages, so learners must infer language-specific generative models on-the-fly to perform accurate sequence modeling.

We begin by providing general background about neural sequence models, ICL and formal languages in \cref{sec:backgroun}, then define the ICLL task in \cref{sec:dataset}. Next, we explore three questions about in-context language learning in neural sequence models:\footnote{Code and reference implementations are released at \url{https://github.com/berlino/seq_icl}}
\begin{mainbox}{Q1: Which model classes can learn to perform ICLL accurately? \hfill (\cref{sec:performance})}
\begin{itemize}[leftmargin=1em]
    \item We find that Transformers significantly outperform recurrent and convolutional LMs at in-context language learning, even when models perform comparably on other model ICL problems.
\item Models with efficient convolutional parameterizations perform especially poorly on ICLL tasks.
\end{itemize}        
\end{mainbox}

\begin{mainbox}{Q2: What algorithmic solutions and circuits do successful in-context language learners implement? \hfill (\cref{sec:interpret})} 
\begin{itemize}[leftmargin=1em]
    \item
    Transformer predictions on ICLL with regular languages are well approximated by smoothed n-gram models.
    \item Transformers trained for regular ICLL develop ``n-gram heads'', higher-order variants of induction heads previously described in LMs \citep{olsson2022context}.
    \item Compared to other model architectures, Transformers better encode in-context n-gram counts in their hidden representations.
\end{itemize}   
\end{mainbox}

\begin{mainbox}{Q3: Can we improve neural models using our understanding of how they perform ICLL? \hfill (\cref{sec:progress})} 
\begin{itemize}[leftmargin=1em]
 \item
    Hard-wiring RNNs and convolutional models with n-gram heads improves their performance on ICLL. 
\item
    These heads are not just useful for ICLL: when equipped with n-gram heads, neural sequence models of all classes exhibit perplexity improvements of up to 6.7\% on natural language modeling tasks.
\end{itemize} 
\end{mainbox}

Our results highlights the usefulness of ICLL as a model problem---not only as a tool for research on ICL, but as a source of insight about architectural features that can improve language modeling in the real world. Many aspects of ICLL, even with regular languages, remain to be understood (e.g.\ learning dynamics and out-of-distribution generalization). Beyond these, ICLL can be extended to even more expressive languages (e.g.\ context-free or context-sensitive languages), offering a path toward understanding of more complex ICL behaviors in real models. Finally, this paper contributes to a growing body of evidence \citep{akyurek2022learning,von2023transformers} that some instances of ICL are best understood as LMs simulating smaller models (here, n-gram language models) using known parameter estimation and inference algorithms.

\section{Background}\label{sec:backgroun}
\subsection{Neural sequence modeling}
Much of modern natural machine learning for natural language processing is concerned with building general-purpose tools for sequence prediction, in which we wish to place a distribution over strings $\vx$. Very often this is done via a product of conditional distributions over \textbf{tokens}:
\begin{equation}
\label{eq:lm}
    p(\vx) = \prod_i p(x_i \mid \vx_{<i}) ~ .
\end{equation}
In practice the distribution $p(x_i \mid \vx_{<i})$ is typically parameterized as a neural network, in which each input token $x_i$ (a symbol, word piece, or word) is assigned an \textbf{embedding} $\ve_i$, which is used to compute a sequence of \textbf{hidden representations} $\vh_i^{(\ell)}$ (one in each layer $\ell$ of the network). The last of these is ultimately used to compute the distribution over next tokens.
A wide variety of architectural choices are available for computing each $\vh^{(\ell)}$ from $\vh^{(\ell-1)}$. 

\subsubsection{Attentional Networks}
Today, the most widely used architecture for neural sequence modeling is the \textbf{Transformer} \citep{vaswani2017attention}. Hidden representations in Transformers are computed via an attention mechanism: to obtain $\vh_i^{\ell}$, models compute weighted averages of previous-layer hidden representations $h_{<i}^{(\ell-1)}$ followed by a feed-forward layer. Letting $\vx =  \vh^{(\ell-1)}$ and  $\vh =  \vh^{(\ell)}$ for readability, each Transformer layer may be expressed as:
\begin{align}\label{eq:attback}
    \vh^{\prime}_i &= \operatorname{softmax}(\mW_k\vx_{<i} (\mW_q\vx_i)^{\top})\mW_v\vx_{<i} ~ , \\
    \vh &= \operatorname{FFN}(\mW_o \vh') ~ ,
\end{align}
where FFN denotes a feed-forward network.

\subsubsection{Recurrent and Convolutional Networks}
Sequence models other than the attention networks can be characterized as recurrent or/and convolutional:
\begin{align}
\label{eq:rnnupdate}
  \text{recurrent} \qquad  & \vh'_{i} = \sigma(\mA \vh'_{i-1} + \mB \vx_i),  \\
\text{convolution} \qquad   & \vh'_{i} = \vl \ast \vx_{<i}, 
\end{align}
where $\mA$, $\mB$ are recurrence parameters, $\sigma$ is an activation function, $\ast$ is a convolution operator and $\vl$ is a convolution filter. However, many of these architectures can equivalently be computed with multiple forms, as we review in \cref{app:architectures}. Hence, we propose a characterization based on \textit{time-invariance} and \emph{linearity}. Time-invariant networks only have parameters that do not depend on input $\vx$, whereas time-variant networks have input-dependent parameters. As a middle ground between them, \textit{weak time-invariant} networks have a mixture of input-dependent and input-independent parameters.  In addition, we use \textit{linear} and \textit{non-linear} to characterize recurrent models, based on whether there is non-linear dependencies among hidden states (i.e., whether $\sigma$ is non-linear).

\paragraph{Linear Time Invariant (LTI)  Models} 
Linear and time invariant models usually have both recurrent and convolutional forms. 
Concretely, their recurrences are linear and the parameters do not change over time, as in $\vh'_i = \mA \vh'_{i-1} + \mB \vx_i$, where $\mA, \mB$ are learnable matrices and $\sigma$ is an identity function and thus omitted. A representative example is the \textbf{S4} model~(\citealp{gu2021efficiently}, along with its many variants, e.g.,~\citealp{gu2022parameterization, mehta2022long}), which demonstrates competitive performance of such forms on sequence modeling.

Extending the LTI principle, models with weak linear time invariance (WLTI) allow for a time-varying component that can still be captured convolutionally after a suitable transformation. When $\vh'_i = \mA\vh'_{i-1} + \mB(\vx_i) \vx_i$, the $\mB$ is not time-invariant anymore, but if it can be written in the form $\vh'_i = \mA\vh'_{i-1} + \mB\phi(\vx_i)$, it may still be viewed as LTI over the transformed input $\phi(\vx_i)$.  Next, we list two kinds of WLTI networks from the literature.

\paragraph{Linear Attention as Weak Time-Invariant Networks}
Linear attention networks such as \textbf{Linear Transformers} \citep{katharopoulos2020transformers} and \textbf{Retentive Networks} \citep[RetNets,][]{sun2023retentive} can be represented by recurrent dynamics wherein hidden states are updated through the accumulation of key-value outer products, as in $\rmS_i = \lambda \rmS_{i-1} + \vk_i^\intercal \vv_i$, where $\rmS$ denotes 2-D hidden states, and $\vk, \vv$ denotes key/value vectors (as in attentional networks) respectively. Since the input of the recurrence (i.e, $\vk, \vv$) is dependent on $\vx$ and $\lambda$ is fixed, we characterize them as WLTI networks. 
Moreover, this recurrent perspective reveals that the effective capacity of the hidden states scales quadratically with their dimensionality, offering a potential explanation for the performance of these models. 

\paragraph{Weak Linear Time Invariant Convolutional Models}
The \textbf{H3} model \citep{dao2022hungry} combines linear attention with a convolutional S4-like layer, resulting in a more complex recurrent form than linear attention with similar input-dependent construction for input (see \cref{app:h3}).   
The \textbf{RWKV} model \citep{peng2023rwkv}, though not originally presented in a convolutional form, adheres to the LTI principle and can be decomposed into a pair of state space models (SSMs), suggesting a potential for convolutional reformulation (\citealp{gu2023mamba}; see \cref{app:rwkv}). The \textbf{Hyena} model \citep{poli2023hyena} is a fully convolutional model with data-dependent gating, as in $ \vh'_{i} = \phi(\vx_i) (\vl \ast \vx_{<i})$, where $\phi(\vx_i)$ denotes a non-linear mapping of $\vx_i$. We also characterize Hyena as WLTI considering the additional input-dependent mapping $\phi(\vx_i)$, compared to vanilla LTI convolutional models.

\paragraph{Linear Time Variant Models} 
Recent state-of-the-art models, \textbf{Mamba} \citep{gu2023mamba} and \textbf{GLA Transformer} \citep{yang2023gated}, feature fully time-dependent parameterization in both the recurrent and input transformations, as in $\vh'_{i} = \mA (\vx_i) \vh'_{i-1} + \mB (\vx_i) \vx_i$. 
Their time-variant nature allows for a more flexible and adaptive response to the input sequence, potentially more expressive in terms of sequence modeling. However, these models represent a departure from the LTI framework and pose new challenges for efficient training, as they do not conform to convolutional structures. 

\paragraph{Non-Linear Time Variant Models}
Finally, we also include \textbf{LSTM}s, which were the most widely used sequence models in the pre-Transformer era.  LSTMs~\citep{hochreiter1997long} feature a complex mapping ($\sigmoid$)   based on gating and intermedia cell states in the recurrence, apart from input-dependent parameterizations, as in $\vh'_{i} = \sigma\big(\mA(\vx_i) \vh'_{i-1} + \mB(\vx_i) \vx_i\big)$ (see \cref{app:lstm}). Such non-linear time variant models are incompatible with efficient parallel training due to complex dependencies among hidden states. Nevertheless, we include LSTMs to examine their expressivity in comparison to recent sequence models, especially recurrent models. 

\subsection{In-context learning}

One feature that has been observed in all model classes above is their capacity for \textbf{in-context learning}. When trained appropriately, sampling from these models given a context of the form:
\begin{equation}
    p_{\textrm{LM}}(~\cdot \mid [\vd_1, f(\vd_1), \septoken, \vd_2, f(\vd_2), \septoken, \dots, \septoken, \vd_n]),
\end{equation}

yields a high-quality estimate of $f(\vd_n)$. Here $\septoken$ is a delimiter token, each $\vd_i$ is an input datum, and $f(\vd_i)$ is its associated output. In practice, $f(\vd)$ may be stochastic and compositional, and both inputs and outputs may be structured (e.g. natural language strings with multiple tokens). 
In addition to this \emph{function-learning} view of ICL, we may understand it more generally as a problem of learning a context-dependent next-token distribution of the same form as \cref{eq:lm}, for some distribution over strings $p_f$. %
Here, sampling from an LM given a context:
\begin{equation}\label{eq:pf}
p_{\textrm{LM}}(~\cdot \mid [\underbrace{x_{1,1}, \ldots, x_{1,n}}_{\vd_1 \sim p_f}, \septoken, \underbrace{x_{2,1}, \ldots, x_{2,n}}_{\vd_2 \sim p_f}, \septoken, \ldots, \septoken, x_{m,1}, \ldots, x_{m,n-1})]),
\end{equation}
yields an estimate of $x_{m,n}$. By chaining these in-context conditional distributions together, we may sample from $p_f$.

ICL may be used to turn general-purpose LMs trained  into models for specific tasks (like machine translation or sentiment analysis). This pre-training process imposes strong priors on how in-context demonstrations are interpreted \citep{min2022rethinking}---for example, providing a small number of translated (English, Spanish) sentence pairs in context will induce a pre-trained LM to perform English--Spanish translation, even though the in-context examples do not fully specify the translation process. But in practice models also seem to be capable of some forms of ``real learning'' in context, inferring functions or distributions within a continuous or combinatorial hypothesis class rather than a fixed set of pre-trained skills \citep{wei2023larger}.

While some work on understanding ICL has studied LMs trained on natural text \citep{olsson2022context, dai2023can}, most interpretability-oriented research has instead studied LMs trained to solve model problems such as linear regression, \citep{akyurek2022learning, von2023transformers}, associative recall \citep{dao2022hungry} or few-shot classification \citep{chan2022data}.
Work in this family studies ICL from several perspectives: as task identification \citep{xie2021explanation, min2022rethinking}, string manipulation \citep{olsson2022context}, or a form of learned ``mesa-optimization'' within a trained LM \citep{akyurek2022learning, von2023transformers, dai2023can}. 
But important aspects of ICL in natural models, including stochasticity, structured outputs, are not captured by these model problems, a limitation we aim to address in this paper. Previous research has mostly focused on Transformers alone, with a few exceptions that have benchmarked an extensive set of neural sequence models \citep{lee2023exploring, arora2023zoology}.

\subsection{Formal Languages}

We study ICL in the context of a new family of model problems involving formal language learning. In these problems, models condition not on a sequence of (input, output) pairs, but instead on a collection of strings sampled from a randomly generated language. Related language-learning problems were also studied by \citet{xie2021explanation} and \citet{hahn2023theory}; here we study models' ability to learn \emph{new} languages rather than recognizing languages from a fixed set seen during training. 
Below we briefly discuss key concepts useful for defining the ICLL task.

\paragraph{Languages, Strings, and Automata}
In the context of formal language theory, a \textbf{language} $L$ is defined as a set of strings over a finite alphabet $\Sigma$. A \textbf{probabilistic language} additionally defines an (optionally normalized) distribution $P(x)$ over the strings $x \in L$. An \textbf{automaton} is an abstract machine that defines a procedure for testing membership in some language, or generating strings from that language.

Our experiments focus on \textbf{regular languages}. These are standardly defined as the set of languages recognized by \textbf{deterministic finite automata} (DFAs). A DFA, in turn, is defined by an \textbf{alphabet} $\Sigma$, a set of \textbf{states} $\mathcal{S}$, an \textbf{initial state} $S_0 \in \gS$, a subset of \textbf{accepting states} $\mathcal{S}_a \subseteq \mathcal{S}$, and a state \textbf{transition function} $T: \gS \times \Sigma \to \gS$. 
To determine whether a DFA accepts some string $\vx = x_1 x_2 x_3 \ldots x_n$, we begin in $S_0$, set $S_i = T(S_i, x_i)$, and finally test of $S_n \in \gS_a$.

To extend this definition probabilistic  languages, we may generalize DFAs to probabilistic finite automata (PFAs) by redefining the transition function as distribution $T: \mathcal{S} \times \Sigma \times \mathcal{S} \to [0, 1]$, the initial states as distribution $I: \mathcal{S} \to [0, 1]$, and the accepting states as distribution $A: \mathcal{S} \to [0, 1]$. These new distributions satisfy the constraints: $\Sigma_{S} A(S) = 1$,  $\Sigma_{S} I(S)=1$ and $F(S) + \Sigma_{x, S'} T(S, x, S') = 1 \quad \forall S$. Under mild conditions  $p_{\textrm{PFA}}(x) = \sum_{S_0, \ldots, S_n} \prod_{i=1}^n  I(S^0) T(S^{i-1}, x_i, S^i) A(S_n)$ is a proper distribution, where the sum is over all possible state sequences that lead to $\vx$. 

In this work, we will use simple PFAs with a single initial state, and without any terminal states (sometimes referred to as NFPAs). Then, we can assign probabilities $p_{\textrm{PFA}}(x)=\sum_{s_0, \ldots, s_n} \prod_{i=1}^n T(s^{i-1}, x_i, s^i)$, where the sum is over all possible state sequences starting from the start state that lead to $\vx$. It can be proven that this is a proper distribution for each sequence length, with the additional convenient property that it can be equivalently represented with a hidden Markov model \footnote{Please refer to 
\cite{DUPONT20051349}
for more discussion of the relationship between NFPAs and their equivalance to HMMs.}

\paragraph{Formal languages and deep networks}

A large body of previous work has used formal languages to probe the limits of neural sequence models \citep{elman1990finding, gers2001lstm, bhattamishra2020ability, suzgun2019memory, hewitt2020rnns, finlayson2022makes}. One line of research has aimed to understand the theoretical expressive power of neural these models. Notably, \cite{merrill2019sequential} prove that LSTMs with bounded precision can recognize counter languages more complex than regular languages, whereas simple RNNs can only recognize regular languages. \cite{merrill2023parallelism} show that bounded precision Transformers cannot recognize important classes of formal languages, including some regular languages. 
Somewhat surprisingly, in light of these theoretical results, our experiments will show that Transformers models outperform recurrent models at learning some regular languages in context, even when they are formally incapable of recognizing even single regular languages in general.

In contrast to this past work, the present study focuses not on whether sequence models can be trained to generate or recognize strings in a \emph{fixed} formal language, but instead whether they can be ``meta-trained'' to adapt on the fly to new languages provided in context.
Closest to this goal, the ``associative recall'' task studied by
\citet{dao2022hungry,arora2023zoology} may also be viewed as a special case of in-context language learning for an extremely restricted sub-class of regular languages; we compare the behavior of models on this task and general ICLL in \cref{sec:performance}.
\citet{bhattamishra2023understanding} study the in-context learning of PARITY and CNF, which may also be cast as formal language learning problems.

\section{\ourdataset: A Benchmark Dataset for In-Context Language Learning}\label{sec:dataset}
\ificml
    \begin{figure*}
\else
    \begin{figure}
\fi
    \centering
    \includegraphics[width=\textwidth]{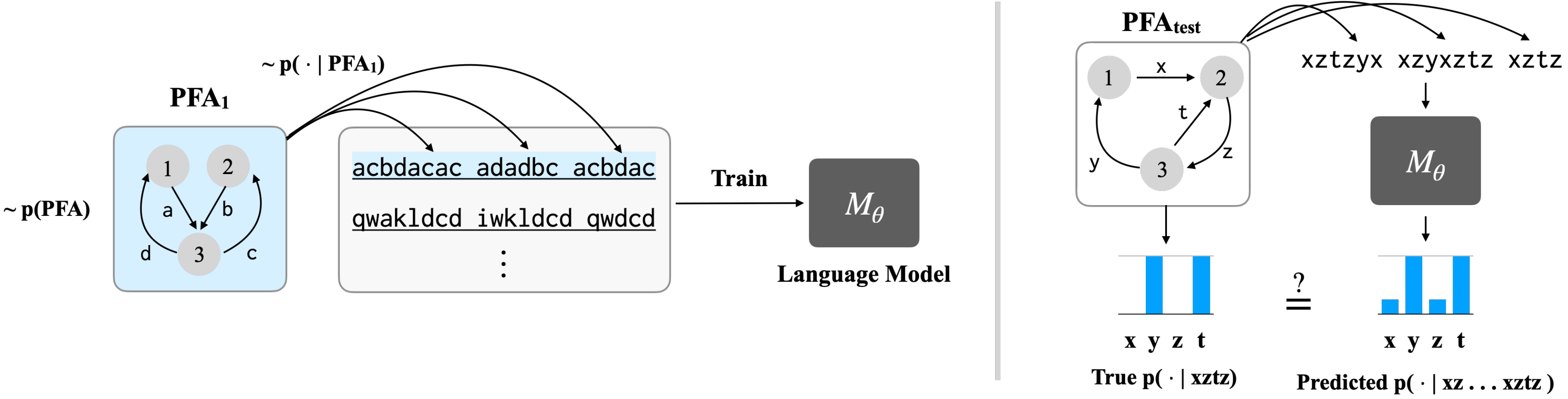}
    \caption{\textbf{ICLL Benchmark:} We randomly generate probabilistic finite automata (PFAs) with uniform transition probabilities, and then generate problem instances that include multiple samples from each PFA. We train and evaluate models on disjoint PFAs for in-context language learning.}
    \label{fig:icllbench}
\ificml
    \end{figure*}
\else
    \end{figure}
\fi

What does it mean to \emph{learn} a formal language of the kind described in \cref{sec:backgroun}? 
Classical formal language theory has studied a number of different learnability criteria, including exact identification \citep{gold1967language},  sometimes with stochastic samples \citep{angluin1988identifying} and probabilistic success criteria \citep{pitt1989probabilistic}.
But if our goal is to understand the behavior of language models, we wish to characterize models' ability to \emph{approximately} predict the next-token distribution given a \emph{finite} set of samples, as in work on PAC learning \citep{valiant1984theory}.
Unlike the PAC setting (but like in natural language modeling) we focus on evaluating ICL given a \emph{fixed} prior distribution over languages.

To do so, we introduce a new dataset called \textbf{\ourdataset}. \ourdataset consists of a set of \textbf{problem instances} $\vd^{(i)}$, each comprising a sequence of \textbf{examples} $[\vd^{(i)}_1, \septoken, \vd^{(i)}_{2}, \septoken, \ldots, \septoken, \vd^{(i)}_n]$, where each example in a problem instance is drawn from the same probabilistic language $L^{(i)}$.
\ourdataset is related to other synthetic language learning datasets, especially the MLRegTest benchmark of \citep{van2023mlregtest}; here we focus on generation rather than membership testing.
To describe how \ourdataset is constructed, we must thus specify (1) how languages are sampled, (2) how strings are sampled from these languages, and (3) how learners are evaluated.

\subsection{Sampling stochastic languages}
\label{sec:samp-lang}

We consider probabilistic languages defined by probabilistic finite automata constructed as follows:

\begin{enumerate}
  \item Sample a number of states $n$ uniformly from the interval $(n_{\min}=4, n_{\max}=12)$. Given this value, define a set of automaton states $\gS = \{S_1, \ldots, S_n\} \cup \{S_0\}$.
  Define the set of accepting states $\gS_a = \{S_1, \ldots, S_n\}$ (excluding $S_0$).
  \item Sample an alphabet size $c$ uniformly from the interval $(c_{\min}=4, c_{\max}=18)$. 
  Sample a language-specific alphabet $V$, containing $c$ symbols, uniformly (without replacement) from a shared symbol set $\mathcal{V}$ (with $|\mathcal{V}| = c_{\max}$).  
  \item For each $S_i$, choose a number of outgoing edges $m_i$ uniformly from $(m_{\min}=1, m_{\max}=4)$. Then, construct a set of edges $(S_i, x_j, S_j)$, where all $x_j$ are sampled uniformly without replacement from $V$, and all $S_j$ are sampled uniformly without replacement from $\{ \mathcal{S}_1, \ldots, S_n\} \setminus S_i$. For every symbol $x'$ not sampled in this step, construct an edge $(S_i, x', S_0)$.
  Together, these edges determine the transition distribution for a (non-probabilistic) DFA $A$.\footnote{This particular choice of transition function ensures that each accepted input corresponds to a unique state sequence. This makes it possible to calculate conditional next token probabilities without needing to marginalize over state sequences}
  \item Construct a new DFA $A'$ by minimizing $A$ \citep{hopcroft1971n}.
  \item Finally, turn $A'$ into a probabilistic automaton without final states by defining each $T(S_i, x_j, S_j) = 1/m_j$ for edges generated above (excluding edges into $S_0$), and $T(S_i, x', S') = 0$ for all other $x', S'$.
\end{enumerate}

This procedure may be run repeatedly to obtain a collection of PFAs $A'$, each with corresponding DFA $A$, and associated with a stochastic language $L$.

\subsection{Sampling strings} 
\label{sec:samp-string}

Given a PFA $A$, sampling from the associated language is straightforward:
\begin{enumerate*}[label=(\textbf{\arabic*})]
    \item Sample a sequence length uniform between $n_{min}=1$ and $n_{max}=50$.
    \item Initialize the sampling procedure with $S_0$.
    \item For each $i \in 1 \dots n$, sample $(x_{i+1}, S_{i+1}) \sim T(S_i, \cdot, \cdot)$.
    \item Return $x_1 \cdots x_n$. 
\end{enumerate*}

\subsection{Dataset}

Using these two sampling procedures, we construct \ourdataset as follows:
\begin{enumerate*}[label=(\textbf{\arabic*})]
    \item Sample a collection of $N_{\text{train}} + N_{\text{test}}$ \emph{distinct} automata $A^{(i)}$ using the procedure in \cref{sec:samp-lang}.
    \item From each automaton, sample $n=10$ to $20$ strings $\vd^{(i)}_j$ with $l=1$ to $50$ symbols each (making the average length of a problem instance  $\approx L = 382$).
    \item Construct problem instances $\vd^{(i)} = [d^{(i)}_1, \septoken, \vd^{(i)}_2, \septoken, \ldots, \septoken, \vd^{(i)}_n]$.
    \item Finally, divide this collection of instances into training and test sets.
\end{enumerate*}

As described below, we use this dataset to evaluate the performance of a variety of neural sequence models on ICLL tasks, while comparing it to other diagnostic tests of ICL.

\section{Which Model Classes Learn to Perform ICLL Efficiently?}\label{sec:performance}
\ificml
    \begin{figure*}
\else
    \begin{figure}
\fi
    \centering
    \includegraphics[width=\textwidth]{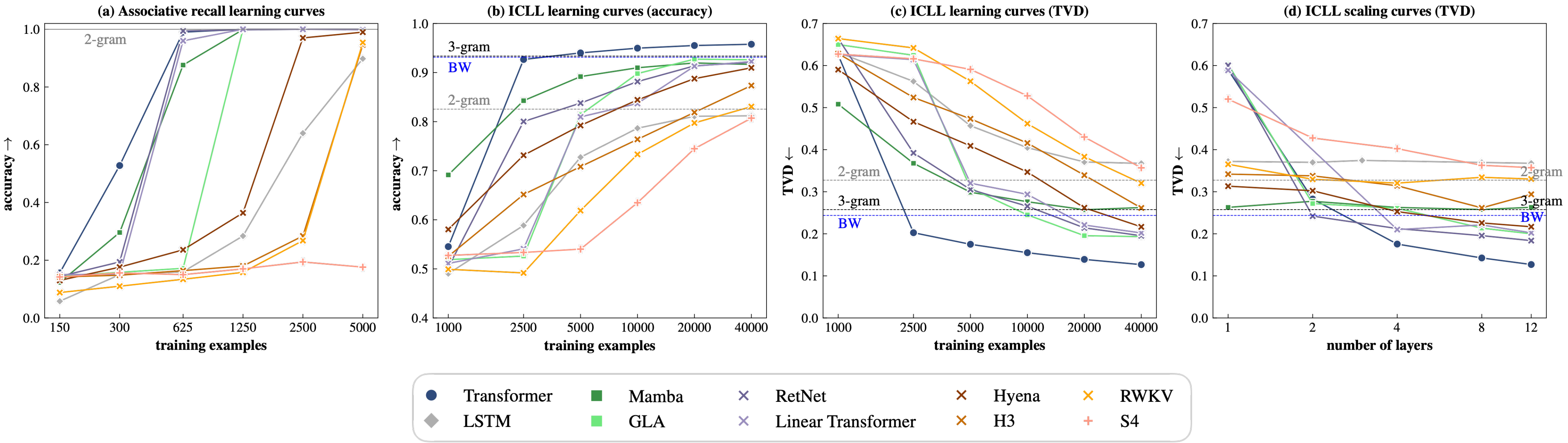}
    \caption{\textbf{{\ourdataset} results:} We propose {\ourdataset} by extending the previous synthetic benchmarks of in-context learning to in-context language learning with regular languages. {\ourdataset} (b, c) yields greater contrast between models comparing to associative recall (a), and enables probabilistical evaluation (c). We find that Transformers are significantly more data efficient recent neural sequence model on in-context learning of regular languages or fine state machines. The Transformer model also adhere a monotonichally increasing scaling curves w.r.t number of layers (d).}

    \label{fig:curves_all}
\ificml
    \end{figure*}
\else
    \end{figure}
\fi

In this section, we use \ourdataset to analyze the behavior of neural sequence models on ICLL tasks. These experiments aim to clarify the relationship between \ourdataset and related existing evaluations of ICL; and most importantly, to determine whether there are meaningful differences between different neural sequence models in their ability to perform ICLL.

\subsection{Setup}
We train models on the \ourdataset dataset to maximize the likelihood:
\begin{equation}
\mathcal{L}(\theta) = \sum_{\vd \, \in \, \mathcal{D}^{\textrm{train}}} \log p_{\theta}(x_{i} \mid x_{<i}) ~ .
\end{equation}

For comparison, we also train models on the associative recall (AR) task introduced by \citet{poli2023hyena}, using a vocabulary size of $\mathcal{V}=40$ (based on \citealp{poli2023hyena}'s ``hard'' setting)  and an input sequence length of $L=382$ (which matches \ourdataset's average sequence length). As with \ourdataset, we use a test set of size 500. For both datasets, we use training subsets of sizes from 150 examples to 40000 examples to evaluate scaling behavior of models.

\subsection{Neural Sequence Models}
We evaluate 10 neural sequence models: Transformers \citep{vaswani2017attention,touvron2023llama}; two Transformer variants with linear attention (RetNet, \citealp{sun2023retentive} and Linear Transformer, \citealp{katharopoulos2020transformers}); four recurrent models (LSTM, \citealp{hochreiter1997long}, RWKV, \citealp{peng2023rwkv}, GLA, \citealp{yang2023gated}, and Mamba, \citealp{gu2023mamba}); and three models with convolutional representations (S4,  \citealp{gu2021efficiently}, H3, \citealp{dao2022hungry}, and Hyena, \citealp{poli2023hyena}).

\subsection{Baseline Learning Algorithms}
To contextualize the performance of these neural models, we also compare to two classical procedures for generative sequence modeling. Given the procedure for sampling languages described in \cref{sec:dataset}, the bayes optimal predictor has the form:
\begin{equation}
    p(x_i \mid x_{<i}) = \sum_{L} p(x_i \mid L) p(L \mid x_{<i}) \propto \sum_{L} p(x_i \mid L) p(x_{<i} \mid L)  p(L),
\end{equation}
where $p(L)$ is the prior that a given language is produced by the sampling process, and $p(\vx \mid L)$ is the probability of an example under $L$. Unlike model problems like linear regression \citep{garg2022can,akyurek2022learning}, there are no general algorithms known for computing this distribution efficiently. However, simple approaches based on maximum likelihood estimation often perform well. Here we evaluate two:
\begin{itemize}
\item \textbf{In-context n-gram models}:
The simplest baselines we consider are \textbf{n-gram} models, which consider a fixed-sized context window of $n-1$ symbols, locate all matching windows within the problem instance, and simply count the number of occurrences of each possible next token across those matches. Our experiments use a variant with backoff \citep{chen1999empirical}---see \cref{app:n-gram} for details.

\item \textbf{In-context DFAs}:
In addition, we compare to a baseline that attempts to explicitly infer the probabilistic automaton that generated the context using the \textbf{Baum--Welch (BW)} algorithm \citep{rabiner1989tutorial}. Given a collection of strings generated by a PFA (or hidden Markov model), this procedure performs maximum likelihood estimation of the transition distribution via Expectation--Maximization \citep{dempster1977maximum}, then performs inference on given context string to perform next-token prediction---see \cref{app:bw} for details.
\end{itemize}

Note that both of these baselines use only the information available \emph{within} an individual problem instance; unlike the learned models, they cannot pool information about the language-generating process across examples.

\subsection{Metrics}\label{sec:metrics}
We evaluate models using two quantities. The first is a greedy-decoding \textbf{accuracy} metric that measures whether each next token predicted by the model is valid under the current regular language:
\begin{equation}
\operatorname{accuracy}(p_{\theta}, L_i) =   \frac{1}{\textrm{NT}} \mathop{\sum}_{\vd^{(i)}}  \mathop{\sum}_{j} {[\mathbf{1}\left[\operatorname{argmax}p_{\theta}(x \mid \vd^{(i)}_{<j}) \in \left\{x': L_i(x' \mid \vd^{(i)}_{<j}) > 0\right\}\right]]},
\end{equation}
where $\textrm{NT}$ is number of total symbols in the test set and $L(x \mid \vd^{(i)}_{<j})$ is the probability of generating $x'$ following the context $\vd^{(i)}_{<j}$ in the language $L$, where this context consists of $j$ tokens comprising of zero or more full examples $\vd$ followed by a partial example.
To provide a finer-grained picture of how well learners have captured the probabilistic aspect of ICLL, we additionally compute
\textbf{total variation distance} between each predicted next-token distribution and the distribution allowed under a given language:
\begin{equation}
\operatorname{tvd}(p_{\theta}, p_{\textrm{pa}})  =  \frac{1}{\textrm{NT}}\mathop{\sum}_{\vd^{(i)}}  \mathop{\sum}_j\left[\frac{1}{2} \sum_{x}\Big| p_{\theta}(x \mid  
 \vd_{<j}^{(i)}) - L(x \mid \vd_{<j}^{(i)})\Big|\right].
\end{equation}

\subsection{Results}

Evaluation results are shown in \cref{fig:curves_all}.

\paragraph{ICLL on \ourdataset shows clear differences across neural sequence models}
On {\ourdataset} (\cref{fig:curves_all}(b, c)), we find that {\tf} models significantly outperform models in all other classes, across both evaluation metrics and a range of training set sizes. Indeed, most non-Transformer models underperform simple n-gram and Baum--Welch baselines, except in the very large data regime.

In contrast, models are less clearly differentiated by the associative recall task (\cref{fig:curves_all}a). Here we observe that all neural models except {\sfour} learn to perform this task with 5000 examples, a training setting used in previous work \citep{dao2022hungry,poli2023hyena}. However, even on this benchmark results are somewhat nuanced---in low data regimes, convolutional models ({\hyena}, {\hthree}, {\sfour}), {\lstm} and {\rwkv} perform comparatively poorly.\footnote{In concurrent work, \cite{arora2023zoology} find that increasing vocabulary size above 1k and querying models with multiple key--value pairs also creates greater separation between models.} 

\paragraph{Deep models are required for ICL, but many models do not benefit from increasing depth}
In \cref{fig:curves_all}, we find that no architecture achieves non-trivial performance on ICLL with only a single layer.  Transformer models monotonically improve their accuracy as the number of layers increases; other models start to overfit to the training set with increasing depth, and do not improve their performance on the test set.

\section{What Algorithmic Solutions do In-Context Language Learners Implement?}\label{sec:interpret}
The previous experiments show that Transformers significantly outperform other neural sequence models at regular ICLL. Can we understand \emph{why} these differences occur? In this section, we analyze the behavior of Transformers trained for ICLL tasks identify the crucial computations they perform, and the features they represent, in the course of ICLL. Our analysis uses three complementary strategies:

\begin{enumerate}[(a)]
\item \textbf{Attention visualization} to interpret the role of individual Transformer sub-components in solving \ourdataset. We find that Transformers develop ``n-gram heads'' that attend to the next token after contexts matching the current context window.

\item \textbf{Probing of hidden representations} to determine where (and how reliably) n-gram statistics are stored. We find that many sequence models learn to encode unigram counts, but Transformers more accurately encode higher-order n-gram statistics.

\item \textbf{Black-box input--output analysis} to evaluate behavioral similarity between Transformers and other learning algorithms known to be effective at ICLL. We find that smaller Transformers are well-approximated by 2-gram models, whereas larger Transformers are closer to 3-gram models. We further find that n-gram models with learned smoothing (obtained by feeding n-gram statistics as input to an MLP-based neural predictor) also perform well at ICLL, and closely matches the behavior of large Transformers.
\end{enumerate}

While attention visualization, black-box analysis, and probing all have limitations as interpretability tools \citep{wen2023transformers,bolukbasi2021interpretability,belinkov2022probing}, the three methods provide convergent evidence that n-gram statistics play a key role in ICLL in transformers.
Each of these findings is discussed in more detail below; in \cref{sec:progress}, we use them to motivate architectural changes that improve the behavior of transformers and other sequence models on both ICLL and natural language modeling tasks.

\subsection{Transformers form in-context n-gram matching heads}
\label{sec:vis}

\paragraph{Setup} We visualize the attention weights of an (8-layer, 1-head) Transformer model on the test sequences from \ourdataset. In \cref{fig:induction_heads}, we plot attention weights between tokens in each layer using heatmaps for an example sequence in the early time steps $(i=50:80)$. Each row shows which tokens the label in that row attends to and the corresponding weights. We display the current DFA state next to each token label on the $y$ and $x$ axes.
\ificml
    \begin{figure*}
\else
    \begin{figure}
\fi
    \centering
    \includegraphics[width=\textwidth]{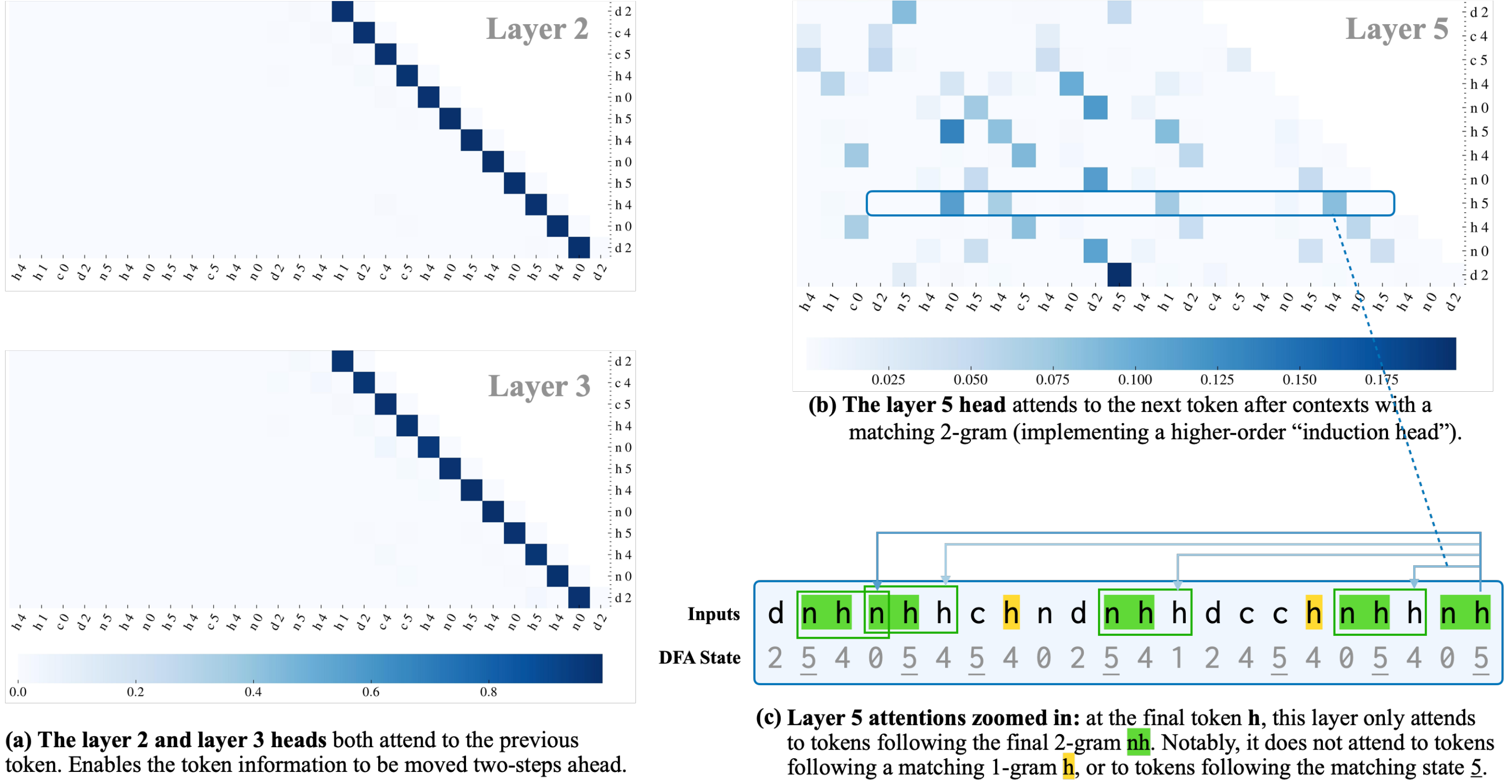}
    \caption{\textbf{N-gram heads in Transformers.} We plot the attention weights of an 8-layer, 1-head Transformer model trained on ICLL with $N=2500$ training examples. Heads in early layers attend to previous tokens (a, b), while the attention head in layer 5 selectively attends to tokens based on 2-gram prefix match rather than the 1-gram match (the original induction head pattern) or the generator DFA's current state.}
    \label{fig:induction_heads}
\ificml
    \end{figure*}
\else
    \end{figure}
\fi

\paragraph{Results} The layer 2 and layer 3 attention heads each attend to the previous token. When composed, these heads enable each hidden representation (starting in layer 3) to incorporate information about the identities of the two tokens that precede it. The layer 5 head then appears to attend to tokens \emph{following} 2-grams matching the most recent 2-gram in the input. In \cref{fig:induction_heads}, for example, the input ends in \texttt{nh}, and the layer 5 head attends to all tokens $X$ in contexts \texttt{nh}$X$. More generally, on inputs of the form $ABC \cdots AB$, we observe that this head attends to tokens in the $C$ position. Notably, these heads do not appear to selectively attend to tokens generated in the same DFA state, as might be expected in a model that is inferring the true data-generating process in context.

The pattern shown in \cref{fig:induction_heads} is a higher-order analog of the ``induction heads'' previously described by \citet{olsson2022context}. Versions of these heads identified in real models have been found to match a single context token; as described by \citet{elhage2021mathematical}, they require a circuit containing a single previous-token head (like our layer 2) followed by a context-matching head (like our layer 5). By composing multiple previous-token heads in series, models are able to match longer contexts.

\subsection{Transformers represent in-context n-gram counts better than other models}

The visualization above shows an attention pattern that is in principle compatible with computation of in-context n-gram statistics, but does not show that this is the quantity these heads in fact compute. In this section, we probe the hidden representations in this model \citep{shi2016does} to determine whether the relevant quantities are encoded by models.

\paragraph{Setup}

To train \textbf{n-gram probes},
we extract the intermediate layer outputs $\vh$ from the models as they process sequences from the training set. For varying values of $n$, we construct an MLP-based probe %
that takes as input a representation $\vh_i$ at time step $i$ and a query token $c$. We train this probe to predict the (unnormalized) \textbf{count} of times $c$ occurs following the same $n-1$ tokens that appear at position $i$ in the input.
We then train similar probes to predict the (normalized) \textbf{frequency} $p(c \mid \vx_{i-n+1}^i) = \frac{\text{count}(\vx_{t-n+1}^ic)}{\text{count}(\vx_{i-n+1}^i )}$ and the binary \textbf{existence} $\textbf{1}[{\text{count}(\vx_{i-n+1}^i c) > 0}]$.

We additionally probe models for whether they encode latent DFA states in their hidden representations. Because the labeling of states is arbitrary, we train these \textbf{state equivalence probes} to take two representations from different timesteps, and predict whether they were generated by the same underlying DFA state.

Implementation details and training hyperparameters for probes are described in more detail in \cref{app:probing}.

\paragraph{Metrics}
We evaluate count and frequency probes according to their relative error
($\frac{|\hat{y} - y | }{y}$; lower is better). We evaluate existence and state equivalence probes according to their binary classification accuracy (higher is better).
We train separate probe models per layer, per model and per task. In \cref{fig:prob25}, we display the result for each task and model at the \textbf{best} layer.

\paragraph{Results}
\ificml
    \begin{figure*}[t]
\else
    \begin{figure}[t]
\fi
    \centering
    \includegraphics[width=\textwidth]{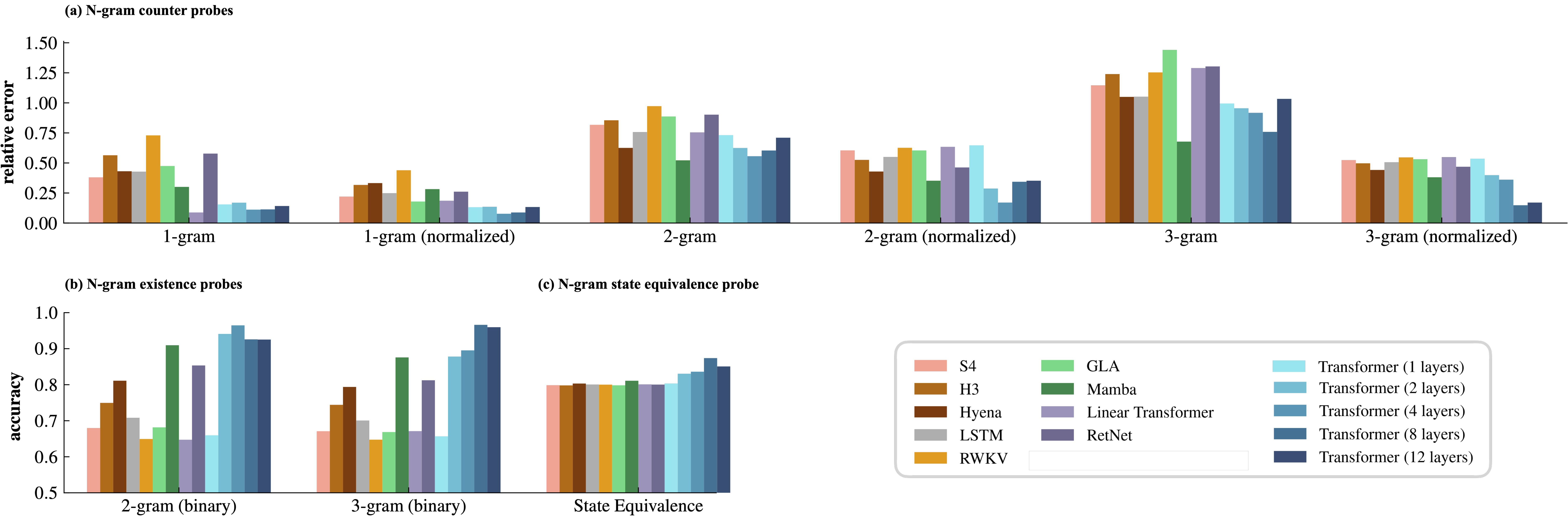}
    \caption{\textbf{Probing analysis of n-gram representations in neural sequence models} (trained with $N_{\textrm{train}}=2500$ examples): (a) Probes are trained to predict the counts and frequencies of the most recent (n-1)-gram + a next query character from the model's hidden state at that time step. Results indicate that Transformer architectures more effectively encode frequencies of higher-order n-grams (bi-grams and tri-grams) compared to other models, with larger Transformer models exhibiting improved performance for higher n-grams. (b) Additional probes assess the ability to detect the presence of an n-gram, revealing that Transformers are superior in this task and in determining state equivalence (c).}
    \label{fig:prob25}
\ificml
    \end{figure*}
\else
    \end{figure}
\fi

The results in \cref{fig:prob25} reveal clear patterns in the information recoverable from models' hidden representations. All models encode normalized unigram frequencies with reasonably low error, though Transformers appear to do so slightly better. For larger context sizes (2-grams and 3-grams), probes on Transformers substantially outperform probes trained on other models---generally the larger the Transformer the better the 3-gram performance.  Interestingly, however, these results do not carry over to unnormalized counts, for which Transformer encodings do not seem to be meaningfully different from other architectures.

In addition to counts, we can an n-gram's presence (\cref{fig:prob25}b) more accurately from Transformers than other models, followed by RetNet and Mamba. More importantly, Transformers hidden states are not just better representative of n-gram features but we can also probe the equivalence of underlying automata states better compared to other model architectures (\cref{fig:prob25}c). We also provide additional results for the models trained with high resource settings in \cref{fig:probe400}, with similar findings. Supplementary results for models trained under high-resource conditions are presented in \cref{fig:probe400}.

\subsection{Transformer predictions resemble n-gram models with learned reweighting}\label{sec:whichalgo}
The previous two sections describe a flow of information through Transformer models
consistent with the hypothesis that Transformer ICLL relies on computation of in-context n-gram statistics. However, it does not explain how models use this information to \emph{predict} a distribution over next tokens. Here, we show that Transformer predictions are also well approximated by those of simpler models with access to the context \emph{only} via n-gram statistics.

\paragraph{Setup}
The experiments in \cref{sec:performance} compared the accuracy of Transformer predictions to standard smoothed 2- and 3-gram models, as well as unsupervised DFA induction. There we observed that Transformers, trained on enough data, more accurately predicted the distribution over next tokens than n-gram language models.
Here, we compare these different models' predictions \emph{to each other}, in addition to the ground-truth language. In doing so, we aim to reveal similarities in prediction strategies across predictors of very different classes. 

In addition to the Transformers, other neural sequence models, and symbolic baselines, we introduce one new model for comparison: a \textbf{learned n-gram reweighting} model. Given an input sequence, we this model first computes a fixed feature representation of this input sequence by computing the empirical next-token distribution in contexts matching the last 1 and 2 tokens of the input, as well as the empirical unigram distribution. It then concatenates these distributions and passes them through an MLP with one hidden layer. This model is trained using the language modeling objective in \cref{eq:lm}, on the same data as other models. We evaluate two variants of this model: one in which the input feature representation contains unnormalized counts of matching n-grams, and another in which the input contains normalized distributions. In automata-theoretic terms, this model may be viewed as a kind of counter automaton \citep{merrill2020linguistic}.

\begin{figure}[t]
    \centering    \includegraphics[height=0.45\textwidth]{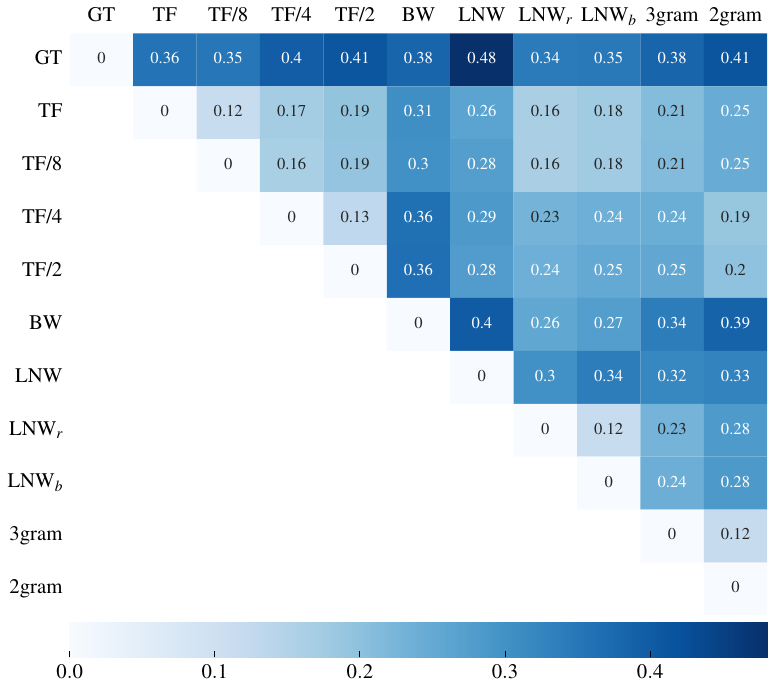}
    \caption{\textbf{Pairwise total variation distance (TVD) over first 100 characters:} We measure total variation distance between pairs of models (trained with $N=2500$ examples) or algorithms across the first 100 tokens of the \ourdataset test set. We find that outputs of the 12-layer Transformer model (TF) are closest to the learned MLP reweighter model with normalized in-context n-gram distributions as input (LNW$_r$). The 3-gram model is the second best match. In the shallow 2-layer Transformer (TF/2) are closer to 2-gram than 3-gram predictions.}
    \label{fig:pairwise}
\end{figure}

These comparisons are shown in \cref{fig:pairwise}.

\paragraph{Results}
Large Transformers with many layers tend to produce next-token distributions more similar to those of n-gram models than to the ground truth DFA or the Baum-Welch algorithm. Specifically, the average total variation distance (TVD) between the large Transformer's predictions and the 3-gram model is 0.21, compared to 0.31 for the Baum-Welch learner (BW), and 0.36 for the next-token distribution in the ground truth language (GT).
Moreover, the learned n-gram reweighting model  (\textsc{LNW}$_r$) best approximates the distributions from the large Transformer, with a TVD of 0.16. 
This suggests the Transformer may implement an implicit smoothing mechanism similar to this learned approach.
Interestingly (and in line with the findings in \cref{sec:vis}) the shallow 2-layer Transformer more closely matches predictions from the 2-gram baseline (TVD of 0.2) than the 3-gram baseline (TVD of 0.25). 

In summary, our experiments show that Transformers compute next token distributions more similar to those of smoothed n-gram models than other neural sequence models.

\section{How Can Findings About ICLL Inform the Design of Neural Sequence Models?}\label{sec:progress}

The previous section suggests that at least part of Transformers' success on ICLL tasks is their ability to efficiently compute in-context n-gram statistics. Can we use this information to improve other models, or to make Transformers more efficient?

The attention pattern associated with ``n-gram heads'' (illustrated in layer 5 of \cref{fig:induction_heads}) may be parameterized as follows:
\begin{equation}
    A(n)_{ij} \propto \mathbf{1}[\underbrace{(\land_{k=1}^n x_{i-k} = x_{j-k-1})}_{\text{n-gram matching}} \land \underbrace{(i > j)}_{\text{causal mask}}],
\end{equation}
where $A_{ij}$ denotes the weight with which the $i$th token attends to the $j$th token.

Building on this observation, we propose to equip models with special \textbf{static n-gram attention heads} (which attend in a fixed fashion, but apply a learned transformation of the attended-to representation) as follows:
\begin{equation}
    \texttt{NGH}^n(h^{(l-1)})_t = W_1\vh^{(l-1)}_t + W_2A(n)^{\top}_t\vh^{(l-1)},
\end{equation}
For a model with a hidden state of size $d$, such a head has $2d^2$ parameters, and only requires two matrix multiplications. To improve a model, we can simply insert it as a standalone new layer between the original layers of an existing model.\footnote{Concurrent work by \cite{arora2023zoology} also inserts sparse attention layers into non-Transformer models from the perspective of efficiency. Their implementation is equivalent to our \texttt{NGH}$^1$, but with learned attention weights; our implementation attends uniformly to all matching contexts.}$^{\text{,}}$\footnote{For Transformer models, we may add these as additional heads in a multi-head attention mechanism.} 

Unlike a standard self-attention layer, $\texttt{NGH}^n$ does not have to store the prefix in memory during model inference. This property makes it compatible with recurrent models.  In particular, in-context ngrams can be compressed via trie (e.g., as in~\citet{pauls2011faster}) and queried efficiently in both time and space. Thus, for models with recurrent form (e.g., GLA/RetNet/Mamba), $\texttt{NGH}^n$ does not introduce much overhead at recurrent inference time.

We evaluate the usefulness of n-gram heads for both ICLL and natural language modeling problems.

\paragraph{Setup: ICLL} Our experiments take an existing architecture (for example RetNet), and insert a sequence of three $\texttt{NGH}$ heads with increasing context size [$\texttt{NGH}^1, \texttt{NGH}^2, \texttt{NGH}^3$] sequentially. We denote this whole bundle as the  \texttt{NGH}$_\textrm{m}^{(1,2,3)}$, where $m$ specifies the original layer after which the $\texttt{NGH}$ heads are added, and the whole architecture as RetNet + \texttt{NGH}$_\textrm{m}^{(1,2,3)}$.

\captionsetup{position=top}

\begin{table}
\centering
\subfloat[{\textbf{n-gram layers bring other models to the Transormer level on ICLL}: We train RetNet and GLA models with n-gram heads on ICLL with $N=2500$ training examples. In TVD metric, adding n-gram layers, brings model performance to the Transformer level trained on the same data without n-gram heads.. In accuracy, hybrid models can outperform Transformer models in the accuracy metric.}]{  
\resizebox{.45\textwidth}{!}{
\begin{tabular}{lll}
\toprule
\multicolumn{1}{c}{\textbf{Model}}  & \multicolumn{1}{c}{\textbf{TVD ($\downarrow$)}} & \multicolumn{1}{c}{\textbf{Accuracy ($\uparrow$)}}\\
\midrule
RetNet \citep{sun2023retentive} & 0.392 & 0.800\\
 \boxSpace\boxRight  \texttt{NGH}$_1^{(1)}$ & 0.310 & 0.814 \\
  \boxSpace\boxRight  \texttt{NGH}$_1^{(1,2)}$ & 0.229 & 0.925 \\
   \boxSpace\boxRight  \texttt{NGH}$_1^{(1,2,3)}$ & 0.217 & 0.94\\
 \midrule
GLA \citep{yang2023gated} & 0.624 & 0.526  \\
 \boxSpace\boxRight  \texttt{NGH}$_1^{(1)}$ & 0.302 & 0.819 \\ 
 \boxSpace\boxRight  \texttt{NGH}$_1^{(1,2)}$ & 0.211 & 0.929 \\ 
 \boxSpace\boxRight  \texttt{NGH}$_1^{(1,2,3)}$ & 0.207 & \textbf{0.946} \\ \midrule
 Transformer \citep{vaswani2017attention} & \textbf{0.203} & 0.926 \\
\bottomrule
\end{tabular}
}} \hfill \subfloat[{\textbf{n-gram layers improves language models}: We train equal sized (340M parameters) language models with and without n-gram heads on 7B tokens from the SlimPajama dataset \citep{cerebras2023slimpajama}. Adding n-gram layers improves each models performance regardless of the model by reducing test set perplexities up to 1.14 point.}]{
\resizebox{.45\textwidth}{!}{
\begin{tabular}{ll}
\toprule
\multicolumn{1}{c}{\textbf{Model}}  & \multicolumn{1}{c}{\textbf{Perplexity ($\downarrow$)}} \\ \midrule
RetNet  \citep{sun2023retentive}           & 16.55        \\
 \boxSpace\boxRight  \texttt{NGH}$_1^{(1,2,3)}$ +  \texttt{NGH}$_{-2}^{(1,2,3)}$  & 15.86 (\textbf{+4.2\%})           \\\midrule
GLA \citep{yang2023gated}                & 15.65    \\
 \boxSpace\boxRight   \texttt{NGH}$_1^{(1,1,1)}$ + \texttt{NGH}$_{-2}^{(1,1,1)}$     & 15.54  (\textbf{+0.7\%})   \\ 
 \boxSpace\boxRight  \texttt{NGH}$_1^{(1,2,3)}$ +  \texttt{NGH}$_{-2}^{(1,2,3)}$     & 15.24  (\textbf{+2.6\%})     \\ \midrule
 Transformer (Llama) \citep{touvron2023llama}      & 16.96         \\
\boxSpace\boxRight \texttt{NGH}$_1^{(1,2,3)}$+  \texttt{NGH}$_{-2}^{(1,2,3)}$    & \textbf{15.82} (\textbf{+6.7\%})        \\
\bottomrule
\end{tabular}
}}
\captionof{table}{\textbf{Hybrid model experiments on ICLL and language modeling.}}
\label{tbl:hybridmodels}%
\end{table}

\paragraph{Results: ICLL}
We experiment by inserting n-gram layers to the base RetNet and GLA models. In \cref{tbl:hybridmodels}(a), the addition of the original induction heads ($\texttt{NGH}^{(1)}$) improves GLA more than 50\% in TVD, and improves RetNet slightly. After adding  the second order induction heads ($\texttt{NGH}^{(1, 2)}$), both RetNet and GLA match Transformer performance. Adding third order induction heads improves the performance even further, enabling the models to outperform the Transformer in the accuracy metric.

\paragraph{Setup: Language Modeling} 
We next  explore whether improvements in ICLL transfer over to language modeling on real data. In these language modeling experiments, we additionally add layer normalization and MLP networks after each \texttt{NGH} layer such that one \texttt{NGH}$_\textrm{m}^{(i)}$ plus MLP makes $4d^2$ parameters, and collectively three \texttt{NGH} layers (i.e., $i \in [1,2,3]$) makes $12d^2$ parameters, matching the number of parameters of a standard Transformer layer.~\footnote{See Appendix for the Python code of \texttt{NGH} layers.} For each base model, we experiment with inserting two n-gram head sequences: one replacing the second layer (\texttt{NGH}$_\textrm{1}^{(1,2,3)}$), and one replacing the second-to-last layer (\texttt{NGH}$_\textrm{-2}^{(1,2,3)}$). 

\paragraph{Results: Language Modeling}
In \cref{tbl:hybridmodels}(b), we show that n-gram heads consistently improve language models. Indeed, the best improvement comes from the Transformer model itself, with a 6.7\% decrease in the perplexity. We also evaluate whether higher-order n-gram heads are really necessary, by evaluating GLA with only 1-gram layers; using all n-grams decreases perplexity 4 times more than using just 1-gram layers (5th row).

The consistent improvements in our setting indicate that n-gram head is one of the mechanisms current sequence models not good at learning from language modeling data itself particularly at 340M parameter scale. Recent work~\citep{arora2023zoology} has shown 1-gram versions of these layers being helpful in synthetic datasets: we show that higher order ($n>1$) n-gram heads extends these improvements to real language modeling in a diverse set of neural sequence models, including Transformers itself.

\section{Conclusion}
In this work, we have proposed a new family of model problems we call in-context language learning (ICLL) that captures important challenges for in-context learning with large language models in real-world settings. Through analysis of model performance on this task, we have identified key differences among classes of neural sequence models, with Transformers emerging as particularly adept at in-context learning. Further investigation has revealed that Transformers succeed by implementing in-context n-gram heads (higher-order induction heads). Inspired by these findings, we demonstrated that inserting simple induction heads for n-gram modeling into neural architectures significantly improves their ICLL performance as well as their language modeling abilities in natural data.
In a broader scope, this paper provides evidence supporting the hypothesis that real language models can in-context learn using known learning algorithms. %

\section*{Acknowledgements}
Thanks to William Merill and Jon Rawski for valuable feedback on an early draft of this paper. Ekin Akyürek and Jacob Andreas are supported by Intel and the National Science Foundation under the PPoSS program (CCF-2217064) as well as the OpenPhilanthropy Foundation. Yoon Kim and Bailin Wang were supported by MIT-IBM Watson AI. 
\bibliographystyle{plainnat}
\bibliography{references}  
\newpage
\appendix
\section{Model Architectures}
\label{app:architectures}
We experiment with a set of neural autoregressive sequence models that model the generative process of a sequence via $ \Pi_{i=1}^T p_{m}\left(\vx_{i+1} \mid \vx_{1:i}\right)$. We characterize each model with three common consecutive modules: (i) an \textbf{embedding layer} that maps input tokens to vectors, (ii) a stack of \textbf{backbone layers}, (iii) and an \textbf{output projection} layer that maps the final outputs from the backbone to the distribution over the token space.
\begin{equation}
p_{m}\left(\vx_{i+1} \mid \vx_{1:i}\right) = \softmax{\left(m_{\textrm{output}} \circ m_{\textrm{backbone}} \circ m_{\textrm{embed}}\left(\vx_{1:i}\right)\right)}
\end{equation}
Collectively, the three modules form a mapping from one-hot vectors to pre-softmax logits, i.e., $\{0, 1\}^{T \times |V|} \rightarrow \R^{T \times |V|}$ where $T$ denotes the sequence length, $|V|$ denotes the vocabulary size.

\paragraph{Embedding Layer ($m_{\textrm{embed}}$)}
The embedding layer is a single projection matrix which converts one-hot input vectors to dense vectors: 
\begin{equation}
    m_{\textrm{embed}}(\vx) = \mW_{e}\,\vx. 
\end{equation}
In almost all models, the positional information need not be incorporated in the embedding layer. It is either explicitly added as the rotary positional embeddings (Rope,~\citet{su2024roformer}) in the attention layer, or implicitly handled in the recurrence/convolution form of models.  The only exception is the original Transformer with multi-head attention, where positional information is added as learnt positional embeddings:
\begin{equation}
    m_{\textrm{embed}}(\vx)_i = \mW_{e}\,\vx_i + \mW_{p}\,\vi,
\end{equation}
where $\vi$ is one-hot representation of the time-step $t$. We use the original Transformer in our synthetic experiments, and use the improved Transformer with Rope in our language modelling experiments on real data.

\paragraph{Backbone Layers ($m^{(l)}_{\textrm{backbone}}$)}
Each backbone layer updates the previous layer's hidden outputs ($\vh^{(l-1)}$) in two sequential steps. First, a \textbf{token mixer} that models the token-level interactions. Effectively,  it can be any \emph{causal} network mapping  hidden states of previous tokens to a new hidden state for the current token:
\begin{equation}
\va^{(l)} = m^{(l)}_{\textrm{mixer}}\left(\vh^{(l-1)}\right),
\end{equation}
where $m^{(l)}_{\textrm{mixer}}\left(\vh^{(l-1)}\right)_i = m^{(l)}_{\textrm{mixer}}\left(\vh^{(l-1)}_{1:i}\right)_t$.
Then, a \textbf{feed-forward} network, $m^{(l)}_{\textrm{FF}}$, applied to the final outputs of the layer with a residual connection:
\begin{equation}
m^{(l)}_{\textrm{backbone}}(\vh^{(l-1)})= m^{(l)}_{\textrm{FF}}\left(\va^{(l)} \right) + \va^{(l)}.
\end{equation}
or alternatively in a gated-linear unit (GAU,~\citet{hua2022transformer}) structure:
\begin{equation}
m^{(l)}_{\textrm{backbone}}(\vh^{(l-1)})= m^{(l)}_{\textrm{GAU}}\left(\vh^{(l-1)}, \va^{(l)}  \right) + \va^{(l)}.
\end{equation}
where $m_{\textrm{GAU}}(\vx, \vy) = \mW_3  (\mW_1 \vx) \odot (\mW_2\vy)$, and $\odot$ denotes element-wise product.~\footnote{Among all the models presented, only Mamba employs the GAU architecture.} In contrast to the token mixer, the feed-forward network is applied individually for each token, i.e., there is no token-wise interactions.

\paragraph{Output Projection ($m_{\textrm{output}}$)} The projection layer consists of a single fully-connected projection that converts the outputs of the backbone $\vh^{(L)}$ to the output space:
\begin{align}
    m_{\textrm{output}}(\vh^{(L)}) = \rmW_{o}\,\vh^{(L)}
\end{align}

In the following sections, we review the model architectures studied in this work - they share the common skeleton presented above, and mainly differ in the design of the token-mixing module $m_{\textrm{mixer}}$.~\footnote{For brevity, we omit the normalization layers which are applied before each token mixer and feed-forward layer.}. We intend to present a unified view of all models by using shared notations and equivalent forms (when possible). 

We will denote input/output of a token mixer as $\vx \in \R^{L \times d}$ and $\vy \in \R^{L \times d}$, respectively. When possible, we will present all the possible forms (i.e., attention-style form, recurrent form and convolutional form) of a model. Generally, attention-style/convolutional forms are useful for developing training schema, whereas recurrent forms are crucial for model's inference schema.

\subsection[Transformers with Self Attention]{Transformers with Self Attention \citep{vaswani2017attention}}\label{app:transformers}
Standard self-attention performs token mixing in the following way:
\begin{align}
    & \vq_i, \vk_j  = \mW_q\vx_i ,\mW_k \vx_j  \in \R^{d_k}   \\
    & \mA_{ij}  \propto \exp(\langle \vq_i, \vk_j \rangle) \in (0,1) \qquad \text{softmax attention} \\
    & \vv_j = \mW_v \vx_j \in \R^{d_v} \\
    & \vz_{i} = \sum_{j=1}^i \mA_{ij} \vv_j \in \R^{d_v} \label{eq:self-att}  \\
    & \vy_{i} = \mW_o \vz_{i} \in \R^{d}
\end{align}
where $d_k, d_v$ denote the dimension for query/key and value vectors, respectively. 
The attention scores are computed based on the pairwise dot-product between the query vector of the current token and key vectors from the context. In the multi-head attention, attention output $\vz_i$ is independently computed  in each head; all outputs are concatenated as the final attention output, which will then be fed into the output projection $\mW_o$. %

\subsection[Transformers with Linear Attention]{Transformers with Linear Attention~\citep{katharopoulos2020transformers}}\label{app:lineartransformers}

The linear attention~\citep{katharopoulos2020transformers} simplifies the standard attention by replacing $\exp(\langle \vq_i, \vk_j \rangle)$ with a kernel map $k(\vq_i, \vk_j )$ with an associative feature map (i.e., $k(\vq_i, \vk_j ) = \phi(\vq_i) \phi(\vk_j)$). In this work, we consider a simple feature map of identity function (i.e., $\phi(\vq_i) = \vq_i$), which yields surprisingly good performance for language model on real data in recent works~\citep{qin2022devil,yang2023gated}. With this feature map, the token mixing process is very similar to standard attention, except that attention scores are not normalized.%
\begin{align}
& \vq_i, \vk_j =  \mW_q\vx_i ,\mW_k \vx_j  \in \R^{d_k}  \\
& \mA_{ij} = \langle \vq_i, \vk_j \rangle \qquad \text{linear attention}   \\
& \vv_j = \mW_v \vx_j \in \R^{d_v}  \\
& \vz_{i} = \sum_{j=1}^i \mA_{ij} \vv_j \in \R^{d_v} \qquad \label{eq:self-att-linear} \\
&  \vy_{i} = \mW_o \vz_{i} \in \R^{d}
\end{align}

The linear attention has an equivalent recurrent form as follows.
\begin{align}
& \vq_i, \vk_j =  \mW_q\vx_i ,\mW_k \vx_j  \in \R^{d_k}  \\
& \rmS_{i} = \rmS_{i-1} + \vk_i^\intercal \vv_i \in \R^{d_k \times d_v} \\
&\vz_i = \vq_i^{\intercal} \rmS_{i} \in \R^{d_v} \\
&\text{(rest is the same as the attention form)}  \nonumber
\end{align}
where $\rmS$ is the 2-D hidden states of the linear recurrence.

\subsection[RetNet]{RetNet~\citep{sun2023retentive,qin2022devil}}\label{app:retnet}
Based on linear attention, RetNet~\footnote{TransNormer proposed in \citet{qin2022devil} has almost the same architecture as RetNet. } further incorporates rotary positional embeddings~\citep{su2024roformer} and a fixed decay rate $\lambda$. The resulting token mixer, namely \textit{retention}, has the following form. 
\begin{align}
    & \vq_i, \vk_j = \mW_q\vx_i,\mW_k \vx_j  \in \R^{d_k} \\
    & \tilde{\vq}_i, \tilde{\vk}_j = \textrm{RoPE}(\vq_{1:i}), \textrm{RoPE}(\vk_{1:j}) \in \R^{d_k}  \\
    &\mA_{ij} = \lambda^{i-j} \langle \tilde \vq_i,  \tilde \vk_j \rangle \\
    & \vv_j = \mW_v \vx_j \in \R^{d_v}  \\
    & \vz_i = \textrm{retention}(\vx_{1:i}) = \sum_{j=1}^i \mA_{ij} \vv_j \in \R^{d_v} \label{eq:self-att-retnet} \\
    & \vr_i = \mW_r\vx_i \in \R^{d_v} \\
    & \vy_{i} = \mW_o \big( \swish(\vr_i) \odot \vz_{i} \big) \in \R^{d}
\end{align}
While the addition of rotary positional embedding to query/key vectors is straightforward in this attention-style form, the additional decay term $\lambda$ is easier to understand in this equivalent recurrent form.
The linear attention has an equivalent recurrent form as follows.
\begin{align}
& \vq_i, \vk_j = \mW_q\vx_i,\mW_k \vx_j  \in \R^{d_k} \\
& \tilde{\vq}_i, \tilde{\vk}_j = \textrm{RoPE}(\vq_{1:i}), \textrm{RoPE}(\vk_{1:j}) \in \R^{d_k}  \\
& \rmS_{i} = \lambda \rmS_{i-1} +  \tilde \vk_i^\intercal \vv_i \in \R^{d_k \times d_v} \\
&\vz_i = \tilde\vq_i^\intercal \rmS_{i} \in \R^{d_v} \\
 & \text{(rest is the same as the attention form)} \nonumber
\end{align}

\subsection[LSTM]{LSTM~\citep{hochreiter1997long}}\label{app:lstm}
All the recurrences we presented are linear in that the there are no non-linear dependencies between adjacent hidden states, e.g., $\partial \rmS_i / \partial \rmS_{i-1}$ is not a function of $\rmS_{i-1}$. For completeness of recurrences, we also consider LSTM \citep{hochreiter1997long}, which is widely used in the pre-Transformer era.  LSTM uses the following non-linear recurrence,
\begin{align}
   \vf_i &= \sigma (\rmW_f \vx_i + \rmU_f \vh_{i-1}  ) \in \R^{d} \\
   \vi_i &= \sigma (\rmW_i \vx_i + \rmU_i \vh_{i-1} )  \in \R^{d} \\
   \vo_i &= \sigma (\rmW_o \vx_i + \rmU_o \vh_{i-1} )  \in \R^{d} \\
   \tilde \vc_i &= \tanh(\rmW_c \vx_i + \rmU_c \vh_{i-1} ) \in \R^{d} \\
   \vc_i & = \vf_i \odot \vc_{i-1} + \vi_i \odot \tilde \vc_i \in \R^{d}  \\
   \vy_i &= \vo_i \odot \tanh(\vc_i) \in \R^{d} 
\end{align}
where $\vf, \vi, \vo$ denotes forget, input and output gate, respectively. To strictly follow the architecture of traditional multi-layer LSTM, we do not use the feed-forward in-between LSTM layers, i.e., the input of  layer $l$ is directly the output from layer $l-1$.
\subsection[GLA]{GLA~\citep{yang2023gated}}\label{app:gla}

Compared with RetNet, GLA incorporated more fine-grained data-dependent gating. Instead of  using the rotary positional embedding, the fine-grained gates can implicitly capture positional information. For the ease of understanding, we first show the recurrent form of GLA, and then its attention-style form.

For each token, GLA additionally relies on two data dependent decay vectors $\balpha_i \in \R^{d_k}$ and $\bbeta_i \in \R^{d_v}$. The outer-product of them (i.e., $\balpha_i^\intercal \bbeta_i $) decides how much information to preserve from previous hidden state $\rmS_{i-1}$.
\begin{align}
& \vq_i, \vk_j  = \mW_q\vx_i, \mW_k \vx_j  \in \R^{d_k} \\
& \balpha_i = \sigma(\mW_\alpha  \vx_i) \in \R^{d_k} \quad \bbeta_j = \sigma(\mW_\bbeta\vx_j) \in \R^{d_v} \\
& \vv_i = \mW_v \vx_i \in \R^{d_v} \\
& \rmS_{i} = \balpha_i^\intercal \bbeta_i \odot \rmS_{i-1} +  \vk_i^\intercal \vv_i \in \R^{d_k \times d_v} \\
& \vz_i = \vq_i^\intercal \rmS_{i} \in \R^{d_v} \\
& \vr_i = \mW_r\vx_i \in \R^{d_v}   \\
& \vy_{i} = \mW_o \big( \swish(\vr_i) \odot \vz_{i} \big) \in \R^{d}
\end{align}
Like linear attention and RetNet, GLA also has the following attention-style form.~\footnote{Please refer to the original paper for the derivation.}  
\begin{align}
    & \vq_i, \vk_j  = \mW_q\vx_i, \mW_k \vx_j  \in \R^{d_k} \\
    & \balpha_i = \sigma(\mW_\alpha  \vx_i) \in \R^{d_k} \quad \bbeta_j = \sigma(\mW_\bbeta\vx_j) \in \R^{d_v} \\
    & \va_i = \prod_i\balpha_{1:i} \in \R^{d_k} \quad \vb_j = \prod_j\bbeta_{1:j} \in \R^{d_v} \\
    & \vv_j = \mW_v \vx_j \in \R^{d_v} \\
    & \tilde \vq_i = \vq_i \odot \va_i  \in \R^{d_k}  \quad \tilde \vk_j = \vk_j / \va_j \in \R^{d_k}  \quad \tilde \vv_j = \vv_j \odot \vb_j \in \R^{d_v}  \\
    & \vz_{i} = \textrm{gla}(\vx_{1:i}) = \big( \sum_{j=1}^i \mA_{ij} \vv_j\big) / \vb_i   \in \R^{d_v} \\
    & \text{(rest is the same as the recurrent form)}  \nonumber
\end{align}

$\odot$ and $/$ denotes element-wise multiplication and division; $\sigma$ denotes a sigmoid function.

\paragraph{Connections among Linear Attention, RetNet, GLA} 
RetNet and GLA both inherit the basic linear recurrence with 2-D hidden states from linear attention. GLA and RetNet mainly differ in the decaying term from the perspective of the recurrent form. Specifically, GLA incorporates fine-grained data-dependent gates ($\balpha, \bbeta$) whereas RetNet uses a single fixed decay $\lambda$ that is shared across all tokens and hidden dimensions. The simplicity of decay in RetNet leads to neat attention-style form needed for parallel training. In comparison, GLA's attention-style form is more nuanced, thus resulting in a more complex training schema. Moreover, RetNet and GLA incorporate the additional output gate $\vr_i$ before the output projection $\mW_o$. Such output gating is also used in LSTM and RWKV models presented below.

\subsection[RWKV]{RWKV~\citep{peng2023rwkv}}\label{app:rwkv}
The recurrence of RWKV is motivated by attention-free network \citep{zhai2021attention}, and it uses 1-D hidden state compared with the recurrences of linear attention.  
\begin{align}
    &  \vk_i = \mW_k \vx_i \in \R^{d_v} \quad \vv_i = \mW_v \vx_i  \in \R^{d_v} \\
    & \va_i = \exp(-\vw) \odot \va_{i- 1} + \exp(\vk_i) \odot \vv_i \in \R^{d_v} \\
    & \vb_i = \exp(-\vw) \odot \vb_{i - 1} +  \exp(\vk_i)  \in \R^{d_v} \\
    & \vz_i = \textrm{wkv}(\vx_{1:i}) = \frac{\va_{i-1} + \exp(\vk_i + \vu)\odot \vv_i}{\vb_{i-1} + \exp(\vk_i + \vu)}  \in \R^{d_v} \\
    & \vr_i =  \mW_r\vx_i \in \R^{d_v}  \\
    & \vy_{i} = \mW_o \big( \sigma(\vr_i) \odot \vz_i ) \in \R^{d_v} 
\end{align}
where $\vw, \vu \in \R^{d_v}$ are learnable parameters, $\sigma$ is an activation function. The $\textrm{WKV}$ operators maintain a recurrence with a pair of states ($\va_i, \vb_i$). Different from linear attention where the 2D hidden state is constructed via an outer-product $\vk_i^{\intercal} \vv_i$, $\textrm{WKV}$ uses element-wise dot-product $\exp(\vk_i) \odot \vv_i$, thus the shape of the key and value vectors are the same.

Since the decay term $\vw$ is not data-dependent, $\textrm{WKV}$ also has the following equivalent convolutional form:
\begin{align}
    & \vk_i = \mW_k \vx_i \in \R^{d_v} \quad \vv_i = \mW_v \vx_i  \in \R^{d_v} \\
    & \tilde{\vkv}_i =  \exp(\vk_i) \odot \vv_i \in \R^{d_v}  \quad \tilde \vk_i =  \exp(\vk_i) \in \R^{d_v}  \\
    & \vl_i = \exp(-i\vw)   \in \R^{d_v}  \\
     & \va = \vl \ast \tilde{\vkv} \in \R^{L \times d_v}  \\
     & \vb = \vl \ast \tilde \vk \in \R^{L \times d_v} \\
     & \text{(rest is the same as the recurrent form)}   \nonumber
\end{align}
where $\ast$ denotes batched long convolution operator, i.e., one dimension of the filter $\vh[:, i] \in \R^{L \times 1}$ handles one corresponding dimension $\va[:, i], \vb[:, i] \in \R^{L \times 1}$.
 
\subsection[S4]{S4~\citep{gu2021efficiently}}\label{app:s4}

Structured state space models (S4) is a family of sequence models defined with four parameters $(\Delta, \rmA, \rmB, \rmC)$. S4 is typically represented as a sequence mapping of $\R^{L \times 1} \rightarrow \R^{L \times 1}$, wherein the input and output are both scalars (i.e. $\vx, \vy \in \R^{1}$). In this case, S4 has the following recurrent form.
\begin{align}
&\vh_i = \bar \rmA \vh_{i-1} + \bar \rmB \vx_i \in \R^{d_k} \\ 
&\vy_i = \rmC \vh_i \in \R^{1} 
\end{align}
where $d_{\textrm{inner}}$ denotes the dimension of hidden states $\vh_i$, $\bar \rmA \in \R^{d_{\textrm{inner}} \times d_{\textrm{inner}}}$ and $\bar \rmB \in \R^{d_{\textrm{inner}} \times 1}$ are transformed parameters for discrete sequence data according to a certain discretization rule (e.g., zero-order hold). $\rmC \in \R^{1 \times d_{\textrm{inner}}}$. Equivalently, it has the following convolutional form:
\begin{align}
&\bar \rmK = [\rmC\bar\rmB, \rmC\bar\rmA\bar\rmB, \dots \rmC\bar\rmA^{L-1}\bar\rmB] \\    
&\vy = \vx \ast \bar \rmK
\end{align}
where $\ast$ denotes the convolution operator and $\bar \rmK $ denotes the convolution kernel. The convolution form is critical on enabling efficient parallel training via Fast Fourier Transform (FFT) algorithms.

Since the recurrent forms of other models are usually presented with vector input/output (i.e., $\vx_i, \vy_i \in \R^{d}$), we present its equivalent batched recurrent form as follows:
\begin{align}
    &\rmS_i = \bar \rmA \circ \rmS_{i-1} + \bar \rmB \circ \vx_i \in \R^{d \ \times d_{\textrm{inner}}} \\
    &\vy_i = \rmC \circ \rmS_i  \in \R^{d} \label{eq:s4-batched}
\end{align}
where $\rmS_i$ denotes 2-D hidden states, $\rmA \in \R^{d  \times d_{\textrm{inner}} \times  d_{\textrm{inner}}}, \rmB \in \R^{d \times  d_{\textrm{inner}} \times 1}, \rmC \in \R^{d \times 1 \times d_{\textrm{inner}}}$, $\circ$ denotes batched matrix multiplication.~\footnote{To differentiate the size of hidden states. in recurrences, we use $\vh_i$ in the 1-D cases; $\rmS_i$ in the 2-D cases.} In this batched form, $d$ numbers of independent SSM run in parallel, each responsible for a dimension of input $\vx$.

\paragraph{Connections to Linear Attentions}
With this batched form in \cref{eq:s4-batched}, it becomes clear that S4, similar to linear attention, enjoys a large 2-D hidden states for recurrence. We can also draw a rough parallel between ($\bar \rmB$, $\rmC$) in S4 to ($\vq_i, \vk_j$) in linear attention as they handle the input and output for the recurrences, respectively. The parallel reveals the difference between S4 and linear attention. In S4, the input and output mapping is not data-dependent, i.e., ($\bar \rmB$, $\rmC$) does not depend on input $\vx$. In comparison, $\vq_i$ and $\vk_j$ are linear mappings of the input $\vx_i$. 

\subsection[H3]{H3~\citep{dao2022hungry}} \label{app:h3}
H3 is a mixture of state-space models and linear attention. In particular, it employs the outer-product structure from linear attention (i.e., $\vk_i^T \vv_i$) to construct the input of a state-space model. 
\begin{align}
    & \vk_i = \mW_k \vx_i \in \R^{d_k} \quad \vv_i =  \mW_v\vx_i \in \R^{d_v} \\
    & \vx'_i = \vk_i^\intercal \vv_i \in \R^{d_k \times d_v} \\
    &  \rmS_i = \bar \rmA  \odot \rmS_{i-1} + \bar \rmB \odot\vx'_i  \in \R^{d_k \times d_v \times d_{\text{inner}}} \qquad   \\
    & \vz'_i =  \rmC \circ \rmS_i    \in \R^{d_k \times d_v} \\
    & \vq_i = \mW_q\vx_i  \in \R^{d_k} \\
    & \vz_i = \vq_i \vz'_i \in \R^{d_v} \\
    & \vy_i = \vz'_i \mW_o \in \R^{d} 
\end{align}
where $\bar \rmA$, $\bar \rmB$, $\rmC \in \R^{d_{\textrm{inner}}}$ are parameters of the state-space models, $\odot$ denotes element-wise product with broadcasting, $\circ$ denotes batched matrix-vector product. The SSM is diagonally parameterized (i.e., $\bar \rmA$ is a vector) and the original H3 paper additionally uses another shift-SSM \citep{dao2022hungry} to further refine the key vector $\vk_i$.

\subsection[Hyena]{Hyena~\citep{poli2023hyena}}\label{app:hyena}

Hyena is a pure convolutional model that does not have an equivalent recurrent form, unlike S4. 
However, it recursively applies the convolution operator at the sequence level for $N$ times (i.e., order-$N$ Hyena). In practice, $N$ is usually set to be $2$, and the resulting form is as follows. 

\begin{align}
  & \vv^n = \mW_{n}\vx \in \R^{L \times d} \\
  & \vz^{0}=  \vv^0 \in \R^{L \times d} \\
  & \begin{rcases}
       & \vl^n_i = \operatorname{FFN}(\vi) \in \R^{L \times d} \\
       & \vz^n = \vv^{n-1} \odot (\vl^n  \ast \vz^{n-1}) \in \R^{L \times d} 
  \end{rcases} \text{ recursion } n=1 \dots N \label{eq:hyena-conv} \\
  & \vy = \vz^N \in \R^{L \times d}
\end{align}

where $\ast$ denotes batched convolution operator. In practice, the filter is padded to the size of $(2L - 1) \times d$ so that the convolution operator becomes a circular convolution for efficient training using FFT.~\footnote{Please refer to Section 2 and 3 of \citep{poli2023hyena} for details.}
Note that the resulting kernels $\vl^1, \vl^2$ do not depend on input $\vx$, but the convolution output is controlled by the data-dependent gate $\vv^1, \vv^2$ in \cref{eq:hyena-conv}. %

\subsection[Mamba]{Mamba~\citep{gu2023mamba}}\label{app:mamba}
Mamba has the same recurrent form as S4, and uses data-dependent parameterization for $\bar \rmA$, $\bar \rmB$, $\rmC$:
\begin{align}
    & \vv_i = \mW_v \vx_i  \in \R^{d_v} \\
    & \rmS_i = \bar \rmA_i \odot \rmS_{i-1} + \bar \rmB_i \odot \vv_i \in \R^{d_k \times d_v} \quad (\text{$\odot$ with broadcast}) \\
    & \vy_i =  \rmC_i  \rmS_i \in \R^{d_v} 
\end{align}
where $\bar \rmA_i, \bar \rmB_i \in \R^{d_k \times d_v}$, $\rmC_i \in \R^{d_v}$ data-dependently parameterized, i.e., computed based on $\vx_i/\vv_i$. However, due to the data-dependence, this recurrent form no longer has an equivalent convolutional form for efficient training. The original paper handles this issue with customized hardware-efficient training algorithms based on the recurrent form.

\section{Optimization \& Hyperparameter Search}
\begin{table}[ht]
    \centering
    \begin{tabular}{ll}
        \toprule
         \bf Hyper Parameter  & \bf Search \\ \midrule
         hidden size & [64, 128, 256, 512, 1024] \\
         number of layers & [1, 2, 4, 8, 12]\\
         number of heads & [1, 2, 4] \\
         epochs & [200, 400]  \\
         batch size & 32 \\
         optimizer & [AdamW] \\
         \boxSpace\boxRight learning rate & [1e-4, 2.5e-4 ] \\
         \boxSpace\boxRight  weight decay & [0.01, 0.1] \\
         \boxSpace\boxRight $\beta$s &  [(0.9, 0.99)] \\
         scheduler & Cosine Scheduler with Warmup \\
          \boxSpace\boxRight minimum learning rate & 2.5e-5 \\
          \boxSpace\boxRight warm-up start learning rate & 1e-7 \\
          \boxSpace\boxRight warm-up steps & 25000 \\
         \bottomrule
    \end{tabular}
    \caption{Hyper-parameter search space for neural models.}
    \label{tab:hyperparam}
\end{table}
We brute force search over grid of hyper-parameters in \cref{tab:hyperparam} and pick the best setting best on validation set on ICLL and AR seperately. In AR we searched hidden sizes up to 256. In ICLL, we search first upto hidden size of 256, then if the best performing hidden size is 256, we try 512, and then 1024. We also used only best performing weight decay of 0.1 and learning rate of 2.5e-4 in the additional search runs. 

\section{Algorithms}\label{app:algs}
\subsection{In-context N-gram Language Model}\label{app:n-gram}
\begin{figure}[ht]
\centering
\begin{minipage}{0.6\linewidth}
\input{algos/ngram}
\end{minipage}
\end{figure}

In \cref{alg:ngram}, we present an \textbf{in-context} applied n-gram model that incorporates a back-off mechanism \citep{chen1999empirical}. This model differs from the standard n-gram approach that is trained on the training set. Instead, here we train a unique n-gram for each example at each time step to predict the subsequent word. The back-off strategy allows for the assignment of non-zero probabilities to unseen n-grams by utilizing information from lower-order n-grams, as shown in line 14 of \cref{alg:ngram}. For each n-gram context $x_{i-N+1}^{i-1}$, the back-off weight $\beta(x_{i-N+1}^{i-1})$ can be computed as follows:

\begin{equation}
    \beta(x_{i-N+1}^{i-1}) = 1 - \sum_{\{w \mid c(x_{i-N+1}^{i-1}w) > 0\}}  \frac{c^{*}(x_{i-N+1}^{i-1}w)}{c^{*}(x_{i-N+1}^{i-1})}
\end{equation}

In the absence of smoothing, the summation is expected to equal 1, resulting in $\beta$ being 0. Smoothing techniques, such as laplace smoothing, modify the counts and allocate a probability mass for unseen n-grams. Alternatively, by excluding the probability corresponding to the padding token $w$, we can reserve probability mass for back-off without explicit smoothing. This approach is employed in our n-gram model implementation and it worked slightly better than add-one smoothing.

Finally, the back-off weights $\alpha$ calculated by normalizing beta for the lower-order n-gram probabilities for the unseen current n-gram sequences:

\begin{equation}
    \alpha(x_{i-N+1}^{i-1}) = \frac{\beta(x_{i-N+1}^{i-1})}{\mathop{\sum}_{\{w \mid c(x_{i-N+1}^{i-1}w) = 0\}}  P(w \mid x_{i-N+2}^{i-1})}
\end{equation}

It is important to note that the normalization ensures that the probabilities of all potential continuations of a given context sum to one.

\subsection{In-context Baum-Welch HMM Language Model}\label{app:bw}
\begin{figure}[ht]
\centering
\begin{minipage}{0.8\linewidth}
\input{algos/bw}
\end{minipage}
\end{figure}
Given a probabilistic automaton, \texttt{PFA}, from \ourdataset, we can construct a Hidden Markov Model (HMM) that assigns the same probabilities to any given string. The construction can be done as:
\begin{itemize}
    \item For each pair of states $S_i, S_j \in \mathcal{S}$, create a corresponding HMM state $H_{(S_i, S_j)}$.
    \item Define the transition probabilities $A(H_{(S_i, S_j)}, H_{(S_l, S_m)}) \propto 1[j=l] \cdot T_{\texttt{PFA}}(S_i, w, S_j)$, where each character $w$ transitions to a unique state in our \texttt{PFA}s.
    \item Set the emission probabilities $B({H_{(S_i, S_j)}, w}) = 1$ if $T_{\texttt{PFA}}(S_i, w, S_j) > 0$, and $0$ otherwise.
    \item Set initial state probabilities 1 for the start states and 0 for the others: $\pi( H_{(S_i, S_j)}) = 1[i==1]$
    
\end{itemize}

The number of states in the constructed HMM is the square of the number of states in the probabilistic automaton. Therefore, we fit an HMM to the examples in \ourdataset with a maximum of $\textrm{NS} = 12^2 = 144$ states. Algorithm \cref{alg:bw} details the in-context Baum-Welch predictor. We begin by constructing a list of observations from the current prefix $x_{1:{i-1}}$, and then fit an HMM given the global vocabulary $\mathcal{V}$ and number of states $\textrm{NS} = 144$ using an improved Baum-Welch algorithm that is consistent with the structure of probabilistic automata in \ourdataset. We incorporate two pieces of prior information about the dataset:

\subsection[Masking A to enforce state transitions]{Masking $A$ to enforce state transitions}\label{app:mask_A}
In our construction, we assume $A_{H_{(S_i, S_j)}H_{(S_l, S_m)}} = 0$ if $j \neq l$. We enforce this constraint in each iteration by masking the corresponding entries in $A$. Additionally, as our PFA sampling schema in \cref{sec:samp-lang} does not include self-transitions, we set all $A_{H_{(S_i, S_i)}H_{(S_i, S_l)}} = 0$.

\subsection[Masking pi to start at the initial states]{Masking $\pi$ to start at the initial state}\label{app:mask_pi}
All our PFAs have a single start state, which we denote as $H_{(S_0, S_i)}$ for all $i$, without loss of generality. We mask all other initial state probabilities such that $\pi(H_{(S_0, S_i)}) = 0$ for $i \neq 0$.

In our experiments, these masking strategies significantly improved the accuracy of the Baum-Welch algorithm. The results presented are based on this modified BW algorithm.

\section{Learned MLP Reweighting}\label{app:mlpsm}
We provide the training details of MLP n-gram reweighting models' used in \cref{sec:whichalgo}, namely LNW, LNW$_r$, and LNW$_b$. 

\paragraph{Count Features (LNW Model)}
Given a data point $d = x_0 \dots x_i \dots x_l = x_0^{l}$, we first extract n-gram features for each position $i$ and for each n-gram length $n$:
\begin{equation}
\textrm{gram}(i; n) = \left[ \textrm{count}_{x_{1:i}}(x_{i-n}^{i-1} w) - 1 \quad \forall w \in \mathcal{V} \right] \in \mathcal{Z}^{|\mathcal{V}|} 
\end{equation}
The full set of n-gram features is the concatenation of n-gram features for $n=1$ to $n=3$:
\begin{equation}
    \textrm{gram}(i; \leq3) = \textrm{concatenate}(\textrm{gram}(i; 1), \textrm{gram}(i; 2), \textrm{gram}(i; 3)) \in \mathcal{Z}^{3|\mathcal{V}|} 
\end{equation}

We then train a sequence model that takes $\textrm{gram}(i; \leq 3)$ as input and applies a 2-layer MLP with GeLU activation to produce the unnormalized scores for the next token distribution. We use the same language modeling loss as in \cref{eq:lm}. Our hyper-parameters for MLP training are as follows:
\begin{center}
\begin{tabular}{ll}
    \toprule
    \textbf{hyper parameter} & \textbf{value} \\ \midrule
    hidden size & 1024 \\
    epochs & 50 \\
    batch size & 32  \\
    optimizer & Adam \\
    \boxSpace\boxRight learning rate & 1e-3\\
    \boxSpace\boxRight $\beta$s & (0.9, 0.99) \\
    scheduler & reduce on plateu \\
    \boxSpace\boxRight patiance & 5 epochs \\
    \boxSpace\boxRight factor & 0.5 \\ 
    \boxSpace\boxRight minimum learning rate  & 1e-5\\
    \bottomrule
\end{tabular}
\end{center}

\paragraph{Frequency Features (LNW$_r$ Model)}
The frequency features model uses normalized n-gram features $\frac{\textrm{gram}(i; n)}{\sum \textrm{n-gram}(i; n)}$ instead of the raw n-gram features explained above.

\paragraph{Binary Features (LNW$_b$ Model)}
The binary features model uses n-gram existence features, where $\textrm{n-gram}(i; n) = 1$ if the n-gram exists at position $i$ and $0$ otherwise, instead of raw n-gram features.

\section{Probing Experiments}\label{app:probing}
\begin{figure}
    \centering
    \includegraphics[width=\textwidth]{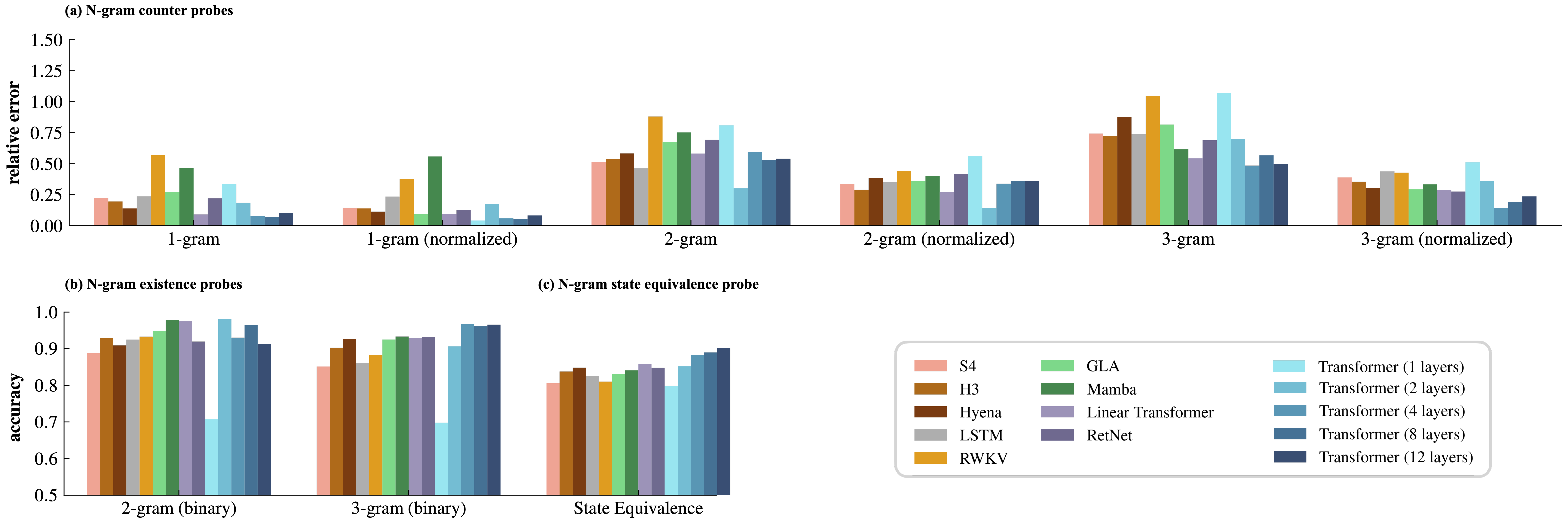}
    \caption{Additional results on probing analysis of n-gram representations with neural sequence models trained with $N_{\textrm{train}}=40000$ examples. See \cref{fig:prob25} for details.}
    \label{fig:probe400}
    \vspace{-6mm}
\end{figure}

\paragraph{Model and Objective}
We train a 2-layer Multilayer Perceptron (MLP) as our probe model in two configurations: \textbf{(1)} $f_{\textrm{n-gram}}(\vh, \vc)$, where $\vh$ represents a hidden state at a specific time step and $\vc$ denotes the query n-gram; and \textbf{(2)} $f_{\textrm{equal}}(\vh_i, \vh_j)$, which is employed in state equivalence probes. The formulation of our $f_{\textrm{n-gram}}(\vh, \vc)$ for an n-gram probe is as follows:
\begin{align}
    \ve_c &=  \mW_{\textrm{embed}}\vc \in \mathbb{R}^{n \times d/2} \\
    \ve_c &= \operatorname{flatten}(\ve_c) \in \mathbb{R}^{nd/2} \\
    \ve_h  &= \mW_{\textrm{proj}}\vh \in \mathbb{R}^{nd/2} \\
    \vx &=\operatorname{concatenate}([\ve_c, \ve_h, \ve_c \odot \ve_h])\\
    \vy &= \mW_2 \operatorname{GeLU}(\mW_1 \vx + \vb_1) + \vb_2
\end{align}
where $n$ is the order of the n-gram and $d$ is the dimensionality of the hidden state used by the probe.
For regression tasks, we employ the mean squared error loss on the output and our targets are corresponding n-gram counts or frequencies. While for classification tasks, we utilize binary cross-entropy loss with the logits being $\vy$, and targets are whether the corresponding n-gram exists or not. 
\paragraph{Data}
We train the probe using hidden states extracted from the actual training set of the models. Specifically, we randomly select an example from the training set, randomly choose a time step within that example, and then create a query n-gram by appending a random next character to the last $n-1$ characters at the chosen time step. For regression tasks, we only consider n-grams that appear at least once in the prefix. Each epoch involves iterating over each example once. For testing, we apply the same sampling procedure using hidden states from the test set.
\paragraph{Model and Objective}
The state equivalence probe $f_{\textrm{equal}}(\vh_i, \vh_j)$ is defined as:
\begin{align}
    \ve_{i}  &= \mW_{\textrm{proj}}\vh_i \in \mathbb{R}^{d} \\
    \ve_{j}  &= \mW_{\textrm{proj}}\vh_j \in \mathbb{R}^{d} \\
    \vx &=\operatorname{concatenate}([\ve_i, \ve_j, \ve_i \odot \ve_j]) \in \mathbb{R}^{3d}\\
    \vy &= \mW_2 \operatorname{GeLU}(\mW_1 \vx + \vb_1) + \vb_2
\end{align}
where $d$ is the dimensionality of the hidden state used by the probe. 
We use the cross-entropy loss in the classification of whether two states are equivalent, and 
$\vy$ serves as the logits for cross-entropy.
\paragraph{Data}
Similar to the n-gram probe, we train the state equivalence probe using hidden states from the model's training set. We randomly select an example and then sample two time steps within it, ensuring that in 50\% of cases the probe receives identical states, and in the remaining 50\%, it receives different states. The testing procedure mirrors that of the training phase.

We employ the following hyperparameters for all probe training. 
We train separate probes for each layer and present the best results in \cref{fig:prob25} and \cref{fig:probe400}.
\begin{center}
\begin{tabular}{ll}
    \toprule
    \textbf{hyper parameter} & \textbf{value} \\ \midrule
    hidden size ($d$) & 128 \\
    epochs & 1000 \\
    batch size & 64  \\
    optimizer & Adam \\
    \boxSpace\boxRight learning rate & 3e-4\\
    \boxSpace\boxRight $\beta$s & (0.9, 0.99) \\
    scheduler & Cosine Annealing  \\
    \boxSpace\boxRight minimum learning rate  & 1e-4\\
    \bottomrule
\end{tabular}
\end{center}

\section{Implementations of N-gram Layers}
In \cref{fig:ngramheadcode} and \cref{fig:ngrammodulecode}, we provide a Python implementation for n-gram layers that we use in our experiments.
\begin{figure}[H]
\begin{python}
def ngram_head(x, hidden_state, shift_step=1, ngram=1):
    """
    Args:
        x: bsz * input_len
        hidden_state: bsz * input_len * d_model
        ngram: 1 means bigram, 2 means trigram
        shift_step: which token to attend to after the matching ngram
    Output:
        bsz * input_len * d_model
    """
    bsz, seq_len = x.shape
    # bsz * L * L, match unigram as the first step
    mask_0 = x[:, None, :] == x[:, :, None]
    causal_mask = torch.tril(torch.ones(seq_len, seq_len, 
        dtype=torch.bool, device=x.device), diagonal=-1 )
    mask_0 = torch.logical_and(mask_0, causal_mask)

    masks = [mask_0.long()]
    for _ in range(1, ngram):
        # mask_0[i, j] = True means token i-1 and token j-1 is matched
        mask_0 = F.pad(mask_0, (1, -1, 1, -1), "constant", False)
        masks.append(mask_0.long())
    ngram_mask = torch.stack(masks, dim=-1).sum(dim=-1) >= ngram
    if shift_step > 0:
        ngram_mask = F.pad(ngram_mask, 
            (shift_step, -shift_step), "constant", False)
    ngram_mask = torch.logical_and(ngram_mask, causal_mask)

    # form a uniform distribution for matched tokens
    ngram_mask_norm = ngram_mask / ngram_mask.sum(dim=2, keepdim=True)
    ngram_mask_norm = torch.nan_to_num(ngram_mask_norm, 0)
    ngram_mask_norm = ngram_mask_norm.to(hidden_state.dtype) 
    output = torch.einsum("bmn,bnz->bmz", ngram_mask_norm, hidden_state)
    return output

class Ngram(nn.Module):
    def __init__(self, d_model, ngram=1):
        super().__init__()
        self.d_model = d_model
        self.ngram = ngram
        self.t0 = nn.Linear(self.d_model, self.d_model)
        self.t1 = nn.Linear(self.d_model, self.d_model)

    def forward(self, x, input_ids):
        bsz, seq_len, _ = x.shape
        h0 = ngram_head(input_ids, x, ngram=self.ngram)
        h1 = x
        y = self.t0(h0) + self.t1(h1)
        return y
\end{python}
\caption{Python implementation of n-gram layers.}
\label{fig:ngramheadcode}
\end{figure}

\begin{figure}[H]
\begin{python}
class NgramBlock(nn.Module):
    def __init__(self, config, ngram):
        """
        Args:
            ngram: 1, 2, or 3
            
        Note: parameter size 4d^2
        """
        super().__init__()
        self.ln_1 = RMSNorm(config.d_model, eps=1e-5)
        
        self.attn = Ngram(config, ngram)
        self.ln_2 = RMSNorm(config.d_model, eps=1e-5)

        mlp_hidden = config.d_model 
        self.mlp = nn.Sequential(
            nn.Linear(config.d_model, mlp_hidden),
            nn.SiLU(),
            nn.Linear(mlp_hidden, config.d_model),
        )

    def forward(self, x, input_ids):
        x_att = self.attn(self.ln_1(x), input_ids)
        x = x + x_att
        x_mlp = self.mlp(self.ln_2(x))
        x = x + x_mlp
        return x
 
\end{python}
\caption{Python implementation of n-gram blocks with SwiGLU-MLP~\citep{shazeer2020glu} and RMSNorm~\citep{zhang2019root}.}
\label{fig:ngrammodulecode}
\end{figure}

\section{Language Model Experiments}
In the language model experiments, all models share the same following training hyperparameters. 
We plan to extend the experiment setting to larger models trained with more tokens in the future.

\begin{center}
\begin{tabular}{ll}
    \toprule
    \textbf{hyper parameter} & \textbf{value} \\ \midrule
    hidden size ($d$) & 1024 \\
    number of training tokens & 7e9 \\
    number of warm-up tokens & 5e8 \\
    batch size (number of tokens) & 5e5   \\
    optimizer & AdamW \\
    weight decay & 0.01 \\
    \boxSpace\boxRight learning rate & 3e-4\\
    \boxSpace\boxRight $\beta$s & (0.9, 0.95) \\
    scheduler & Cosine Annealing  \\
    \boxSpace\boxRight minimum learning rate  & 3e-5\\
    \bottomrule
\end{tabular}
\end{center}

\end{document}